\documentclass[11pt]{article}

\usepackage{amssymb}
\usepackage{graphicx}
\graphicspath{{../figures/}}
\usepackage[final]{acl}
\usepackage{amsmath} 
\usepackage{times}
\usepackage{latexsym}
\usepackage{booktabs}
\usepackage[T1]{fontenc}

\usepackage[utf8]{inputenc}

\usepackage{microtype}

\usepackage{inconsolata}

\usepackage{graphicx}

\newcommand{\lals}{\textsc{Lals}}

\title{Vision-Language Models Suppress Female \\Representations Under Ambiguous Input}
\author{
  \textbf{Arnau Marin-Llobet}\textsuperscript{1} \quad
  \textbf{Simon Henniger}\textsuperscript{1} \quad
  \textbf{Mahzarin R. Banaji}\textsuperscript{2} \\
  \textsuperscript{1}School of Engineering and Applied Sciences, Harvard University \\
  \textsuperscript{2}Department of Psychology, Harvard University \\
  \texttt{amarinllobet@seas.harvard.edu}, \texttt{mahzarin\_banaji@harvard.edu}
}
\begin{document}
\maketitle

\begin{abstract}
Alignment teaches vision-language models (VLMs) to avoid expressing demographic biases, and when gender is clearly visible they largely succeed. Far less is known about ambiguous inputs (a worker in full gear, a figure seen from behind) cases common in practice yet rarely studied. We find that minimal prompting pressure exposes occupation–gender defaults when prompting ambiguous input images, with models collapsing to \emph{male} even for strongly female-stereotyped occupations. But do these outputs reflect what models actually encode internally? We introduce \lals{} (Latent Association Leaning Score), a zero-shot metric that projects visual-token activations into the model's text-embedding space to measure concept associations per token and layer. Across 15 occupations, over 800 gender-ambiguous images, and four VLMs, internal representations and outputs often becomes systematically decoupled: models often encode a female association internally yet output male. Layer-wise analysis reveals an asymmetric filter—male signal amplifies end-to-end while female signal peaks mid-network and is suppressed before generation—and a color ablation shows that culturally loaded visual cues such as clothing color further modulate these internal associations.
\end{abstract}

\section{Introduction}
\label{sec:intro}

\begin{figure}[ht]
  \centering
  \includegraphics[width=0.8\linewidth]{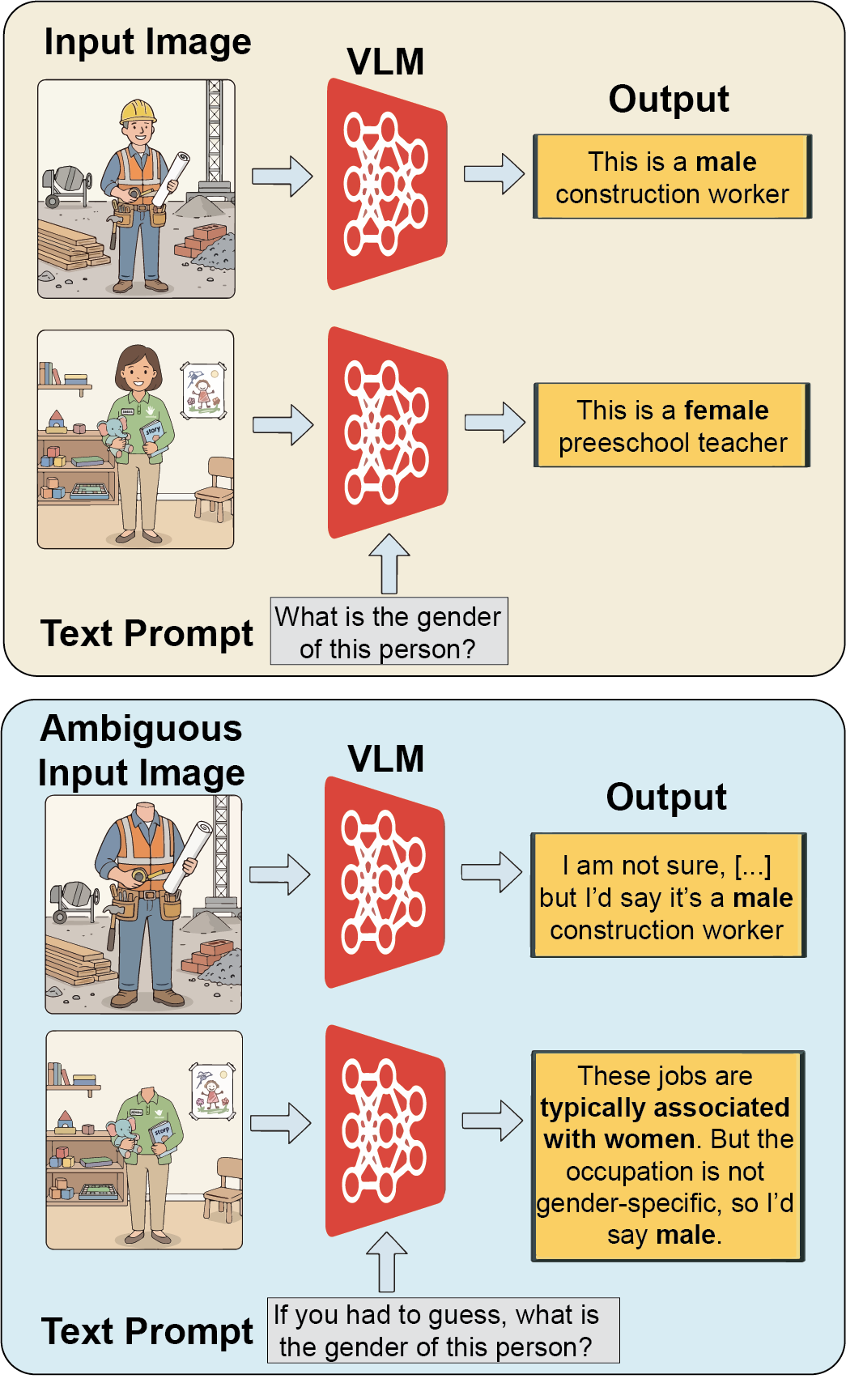}
  \caption{\textbf{Representative Summary of Findings.} \textit{Top:} when gender is visually clear, VLMs report it accurately. \textit{Bottom:} when the image is gender-ambiguous (faceless figures, same occupations), models default to \textit{male} under forced-choice prompting, even for female-stereotyped roles.}
  \label{fig:setup}
\end{figure}

Vision-language models (VLMs) are increasingly used in applications where fairness matters---from content moderation to image retrieval to assistive tools that describe visual scenes.  As these models enter high-stakes settings, auditing them for bias has become a priority.  The standard approach is straightforward: show the model an image, ask it a question, and check whether the output reflects stereotypical or harmful associations.  If a model describes a doctor as ``he'' or a nurse as ``she'' when gender is ambiguous, these types of bias might be flagged \cite{vo2025vision}.

This output-level auditing has driven significant progress. Alignment techniques such as RLHF \cite{ouyang2022training} have made modern VLMs remarkably careful: when asked to describe an image of a worker whose gender is not visible, they generally answer ``a person'' rather than ``a man'' or ``a woman.'' We show that  these outputs are only the surface, and the bias remains underneath.  A model that produces neutral text may still carry biased \emph{representations}---associations encoded in the activations of its visual tokens that shape downstream behavior even if they do not appear in the final response.  These internal associations matter for at least two reasons.  First, VLM embeddings are increasingly used as features for downstream systems (image search, content ranking, hiring tools), where biased representations propagate without ever passing through the model's language thinking process.  Second, output neutrality or clean inputs are a fragile condition: biases suppressed by alignment may resurface under different prompting strategies, not very clear visual inputs, or even fine-tuning, or deployment conditions.

In this paper, we ask a simple question: {do VLMs' internal visual representations carry the same gender associations as their outputs, even when the input images are ambiguous?} To answer this, we introduce \lals{} (Latent Association Leaning Score), a zero-shot metric that measures concept associations at the level of individual visual tokens and layers.  \lals{} builds on recent work showing that visual token activations in VLMs can be projected into the model's text embedding space, enabling a direct reading of what each image patch ``encodes'' at any point in the network \cite{krojer2026latentlens}.  By comparing these decoded representations against a gender-balanced reference corpus, \lals{} produces a continuous score (from male-leaning to female-leaning) for every token at every layer, without any training. Our main findings are:

\begin{enumerate}
\item \textbf{Internal representations and outputs are decoupled when input images are ambiguous.}  We identify three regimes: stereotypical occupations where internals and outputs agree on male (e.g., firefighter), where both agree on female (e.g., makeup artists), and sometimes where models internally encode female associations but output male (e.g., babysitter). This divergence regime represents a concrete blind spot for output-level auditing.

\item \textbf{Late layers act as an asymmetric filter.}  Sweeping \lals{} across layers reveals that male associations amplify from early to late layers, while female associations peak in the mid-late of the network and are suppressed toward the output.  This mechanism might be a potential explanation on why the male default dominates outputs even for occupations that are internally female-associated in non-obvious gender images.

\item \textbf{Internal associations are shaped by culturally loaded visual cues.}  A color ablation shows that changing the clothing from blue to pink substantially reduces the internal male signal---not because the model is confused by color, but probably because it has learned the cultural gender associations that colors carry.
\end{enumerate}
\section{Related Work}
\label{sec:related}
\paragraph{Bias auditing in vision-language models.}
Work on VLM bias has overwhelmingly operated at the output level. Early studies documented gender and racial biases in image captioning~\cite{zhao2017men, burns2018women, tang2021mitigating} and image search~\cite{wang2021gender}, and more recent benchmarks evaluate VLMs on occupation--gender defaults, counterfactual image pairs, and stereotype-consistent prompts~\cite{hall2024visogender, fraser2024examining, janghorbani2023multimodal, howard2024socialcounterfactuals, xiao2025genderbiasvl, girrbach2025revealing}. All of these assume that a model's output is a faithful window into its internal associations. In NLP, this assumption has been challenged: linear probes and embedding-space analyses repeatedly show that demographic biases persist after output-level debiasing~\cite{bolukbasi2016man, caliskan2017semantics, may2019measuring, guo2021detecting, gonen2019lipstick}. Extending this line to VLMs remains underexplored. Most representation-level analyses focus on feature quality rather than social bias~\cite{tong2024eyes}; the few exceptions operate on contrastive encoders rather than generative VLMs~\cite{konavoor2025strong}, localize bias to the image encoder by causal mediation without quantifying what each layer encodes~\cite{zhang2024images}, train sparse autoencoders to isolate and suppress demographic ``social neurons''~\cite{an2026debiaslens}, or, closest to us, track bias in responses and hidden states across layers, finding that fairness peaks mid-network and erodes late~\cite{lan2025unveiling}. This work extends this line: we are zero-shot and operate at token-level granularity, identifying \emph{which image patches} carry biased associations and \emph{how}, and in which direction, they evolve across layers.
\paragraph{Interpreting internal representations in vision models.}
A growing line of work reads intermediate representations by projecting them into interpretable spaces. LogitLens~\cite{nostalgebraist2020logitlens} projects hidden states into the output vocabulary, giving a coarse, word-level reading; TunedLens~\cite{belrose2023eliciting} refines this with learned per-layer affine transforms, in part because raw logit-lens readouts are unreliable in early layers, and Patchscopes~\cite{ghandeharioun2024patchscopes} decodes hidden states through a separate inference pass. Recent work has extended these tools to VLMs: LatentLens~\cite{krojer2026latentlens} shows that visual token activations can be meaningfully projected into the model's text-embedding space, \citet{ma2025interpreting} use a logit lens to trace how object information flows through VLM layers, and \citet{ananthram2025see} apply one over visual tokens to probe cultural bias. \lals{} adapts this projection in a new direction: rather than using it for general-purpose interpretability, we pair it with a structured text reference corpus to quantify demographic associations in a zero-shot, token-level manner. A complementary tradition---activation patching~\cite{meng2022locating} and causal tracing~\cite{vig2020causal,zhang2024images}---identifies which components are causally responsible for a behaviour by intervening on activations. These methods locate \emph{where} a decision is made; our work measures \emph{what} is encoded at each location. Our layer-sweep analysis connects the two by tracing how gender signal propagates through the network.
\section{Approach}
\label{sec:method}

\begin{figure*}[t]
  \centering
  \includegraphics[width=\linewidth]{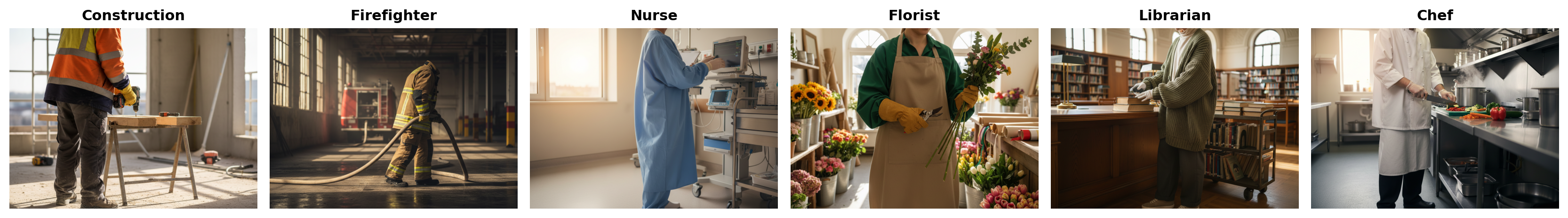}
  \caption{Representative ambiguous-gender images. Each shows
  a faceless figure in an occupation-specific setting with no
  visible cues.}
  \label{fig:ambiguous-examples}
\end{figure*}

\subsection{\lals{}: Latent Association Leaning Score}
\label{sec:lals_metric}

\lals{} measures the degree to which a visual token's internal
representation is associated with one pole of a concept dimension
(e.g., male vs.\ female).  It requires no training and operates at
the level of individual tokens and layers.

\paragraph{Reference corpus.}
We construct two balanced word lists for the target concept.  For gender, one list contains male-associated terms (\emph{man, father, boy, husband, \ldots}) and the other female-associated terms (\emph{woman, mother, girl, wife, \ldots}), including names and role terms.  Each term is embedded using the VLM's own text encoder, producing a reference database $\mathcal{D} = \{(\mathbf{e}_i, g_i)\}$ where $\mathbf{e}_i$ is the text embedding and $g_i \in \{+1, -1\}$ indicates the concept pole.

\paragraph{Visual token projection.}
Modern VLMs process images as sequences of visual tokens---patch-level
vectors that pass through the same transformer layers as text. At any layer $\ell$, we extract each visual token's hidden state $\mathbf{h}_t^\ell$ and project it into the text embedding space using the LatentLens procedure \cite{krojer2026latentlens} , yielding a vector $\mathbf{v}_t^\ell$ that lives in the same space as the reference corpus.  This lets us directly compare what
each image patch encodes against gendered text concepts.

\paragraph{Scoring and aggregation.}
For each projected token, we retrieve its $k$ nearest neighbors
from $\mathcal{D}$ by cosine similarity and compute the gender
balance:
\begin{equation}
  \text{LALS}(t, \ell) \;=\;
    \frac{1}{k} \sum_{i \in \mathcal{N}_k(\mathbf{v}_t^\ell)} g_i
\label{eq:lals_token}
\end{equation}
This produces a score in $[-1, +1]$: fully male-associated, fully female-associated, or balanced. To obtain an image-level score, we aggregate over the top 5\% of tokens by absolute magnitude (validated empirically in Fig. \ref{fig:top_pct_ablation}, appendix), focusing on the patches with the strongest signal:
\begin{equation}
  \text{LALS}_{\text{image}}(\ell) \;=\;
    \frac{1}{|\mathcal{T}_{5\%}|}
    \sum_{t \,\in\, \mathcal{T}_{5\%}}
    \text{LALS}(t, \ell)
\label{eq:lals_image}
\end{equation}
Negative values indicate male-leaning representations, positive
values female-leaning, and values near zero no detectable
association.

\paragraph{Properties.}
\lals{} is \emph{zero-shot} (no labeled images needed),
\emph{token-level} (revealing which image regions carry the
association), \emph{layer-level} (tracing how associations evolve
through the network), and \emph{concept-general} (swapping the
reference corpus audits any attribute expressible as opposing text
poles).

\subsection{Experimental Setup}
\label{sec:setup}

\paragraph{Models.}
We evaluate four open-weight, instruction-tuned VLMs with different
architectures, vision encoders, and vision--language connectors:
Qwen2-VL-7B \cite{Qwen2VL}, Qwen2.5-VL-7B \cite{qwen25}, LLaVA-v1.6-Mistral-7B \cite{liu2023visual}, and InternVL2.5-8B \cite{chen2024expanding}. We report \lals{} with $k = 20$ neighbors and top-5\% aggregation, unless stated otherwise.

\paragraph{Ambiguous-person dataset.}
We use Google Gemini 2.5 Flash (image generation mode)~\cite{comanici2025gemini} to generate images of faceless or obscured figures in occupation-specific settings, where gender cannot be determined from visual cues alone (Figure~\ref{fig:ambiguous-examples}).  A human annotator verified every image, discarding any with visible gender markers. The final dataset spans 15 occupations---male-stereotyped (e.g., firefighter, construction worker), female-stereotyped (e.g., nurse, florist), and neutral (e.g., chef, waiter)---with 60 images per occupation unless stated otherwise.

\paragraph{Output responses.}
To compare internal representations with output behavior, we query each model with two prompt types. \emph{Open-ended:} ``Describe what this person is doing''---testing whether the model spontaneously attributes gender. \emph{Forced-choice (FC):} ``If you had to guess, is this person male or female? Answer in one word''---forcing an explicit commitment.  We also run the FC prompt without any image to measure each model's text-only prior. 

\section{Results}
\label{sec:results}

\subsection{\lals{} Detects Bias Signal}
\label{sec:validation}

Before applying \lals{} to ambiguous images, we verify that the metric (i)~detects genuine gender signal when it is visually present, (ii)~produces no spurious signal when people are absent, and (iii)~is robust to methodological perturbations.

\begin{figure}[h]
\centering
\includegraphics[width=\columnwidth]{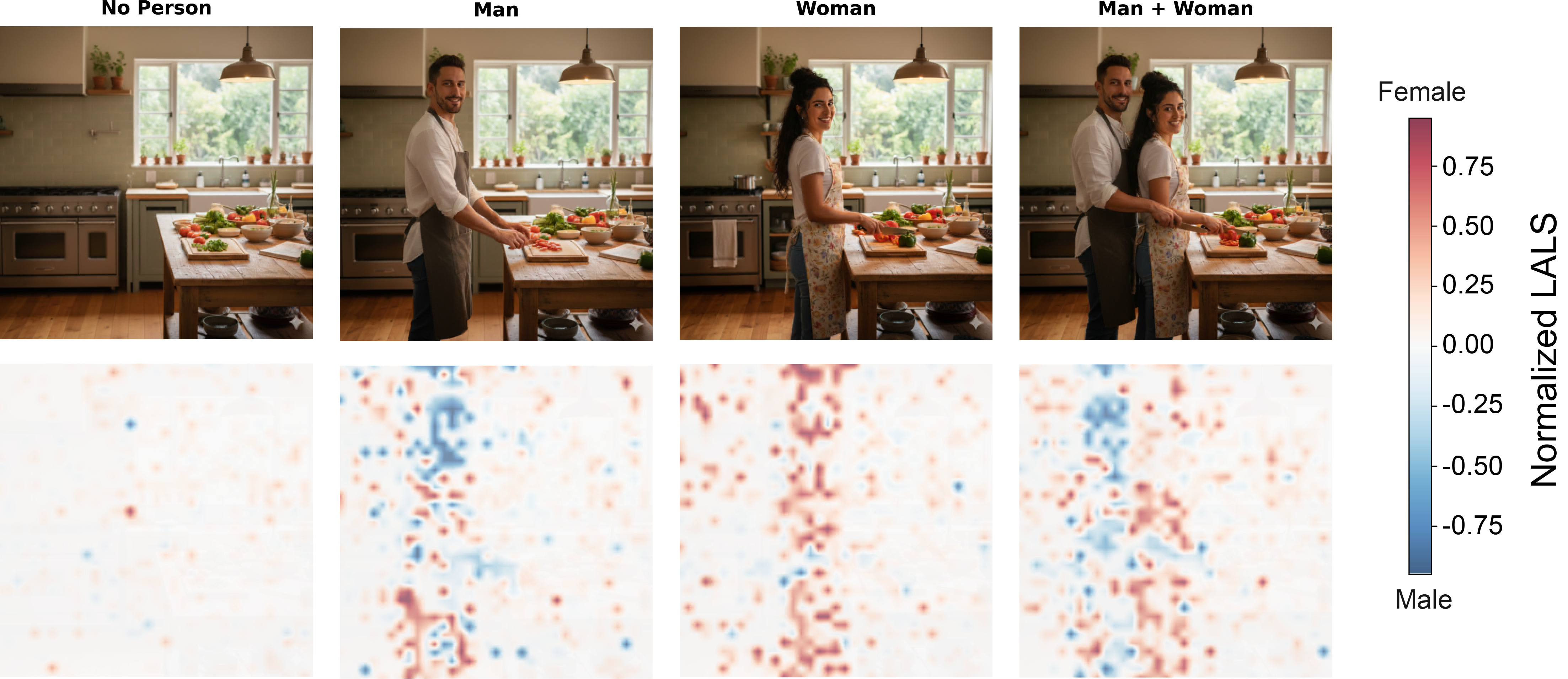}
\caption{\textbf{\lals{} heatmaps for a kitchen scene under four conditions.} \textit{No Person:} near-zero signal throughout. \textit{Man / Woman:} inserting a single person produces a gender-consistent signal localized on the individual. \textit{Man + Woman:} \lals{} correctly assigns male (blue) and female (red) signal to the respective individuals.}
\label{fig:validation_panel}
\end{figure}

\paragraph{Localization on matched scenes.} We construct matched scene sets in which the same background is shown with no person, a man, a woman, or both, isolating \lals{} responses to gender-visible individuals while holding scene context constant. Figure~\ref{fig:validation_panel} illustrates a kitchen scene under all four conditions. With no person present, the heatmap is nearly flat and the net \lals{} hovers near zero. Adding a man produces a clear male-leaning (blue) cluster localized on the person; adding a woman produces the opposite female-leaning (red) pattern in the corresponding region. When both are present, \lals{} correctly assigns male and female signal to the respective individuals. The pattern replicates on a construction-site scene (Appendix Fig.~\ref{fig:working_panel}), and across person-free images ($N{=}10$) all net \lals{} values fall close to zero (mean~$=+0.001$, $\sigma{=}0.005$).

\paragraph{Controls.} Two additional checks guard against artifacts. Randomly permuting the gender labels in the reference corpus (\emph{shuffled database}) collapses the signal by 98\%, confirming that \lals{} depends on correct text--embedding alignment rather than on distributional properties of the embedding space. Varying the neighborhood size ($k \in \{10, 20, 50\}$) produces stable results ($<\!15\%$ variation in gender delta), indicating that \lals{} is not sensitive to the exact number of nearest neighbors.

\paragraph{Cross-check with a supervised probe.} As an independent check, we train a logistic regression probe on visible-gender hidden states ($N{=}200$, 5-fold cross-validation) to predict binary gender from visual token representations. The probe achieves 97\% accuracy at layer~4 and 94.5\% at layer~16. Applied to ambiguous-occupation images, the probe's per-image $P(\text{female})$ correlates with \lals{} ($r=0.52$, $p=0.003$), confirming that both approaches capture overlapping structure in the representations. The moderate rather than near-perfect correlation is expected: the probe learns a single linear boundary, while \lals{} aggregates over a broader neighborhood of the embedding space. The probe is trained on visible-gender images and so relies on overt cues that ambiguous images lack, extrapolating a single linear boundary, whereas \lals{} aggregates contextual cues such as scene, clothing and color. Note that this cross-check, the causal ablation (Appendix~\ref{sec:causal}), and the analysis such as the color ablation are all run on the ambiguous images themselves.

\subsection{Outputs Collapse Toward Male Under Ambiguity}
\label{sec:male_collapse}

To study how gender is represented in ambiguous inputs we first ask what models \emph{say} when shown gender-ambiguous images. When prompted in open-ended format (``Describe what this person is doing''), the Qwen models and InternVL produce gender-neutral responses in essentially every case: ``the person is arranging flowers,'' not ``the woman is arranging flowers,'' or they refuse to attribute gender at all. LLaVA is the exception, spontaneously gendering the figure in a minority of responses---almost always as male (e.g., 80\% of delivery drivers, 57\% of maids/cleaners; Table~\ref{tab:output-behavior}). This is the expected effect of alignment training.

The behavior changes immediately under minimal prompt pressure. With a forced-choice (FC) prompt---``If you had to guess, is this person male or female?''---occupation-dependent defaults emerge sharply (Table~\ref{tab:output-behavior}). Occupations such as firefighters are classified as male in 100\% of images across all four models, which is unsurprising. More strikingly, most female-stereotyped occupations also collapse toward male: hairdresser (92\% BLS female) is classified as male 88--97\% of the time across all four models, babysitter (93\% BLS female) is majority male in all models (72--97\%), and preschool teacher (97\% BLS female) is majority male in two of four models (LLaVA and InternVL). Even nurse---one of the most strongly female-coded occupations in the U.S.\ labor force at 87\% BLS female~\cite{bls2025cps}---is classified as male by LLaVA. The surface neutrality of open-ended outputs masks biases that become visible the moment a model is forced to commit.

\begin{figure}[ht]
  \centering
  \includegraphics[width=\linewidth]{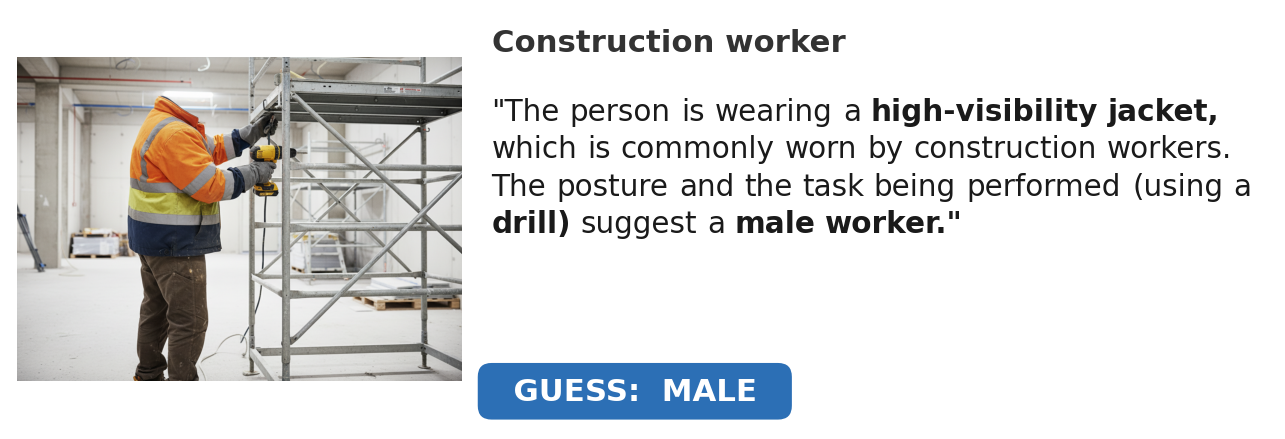}
  \\[4pt]
  \includegraphics[width=\linewidth]{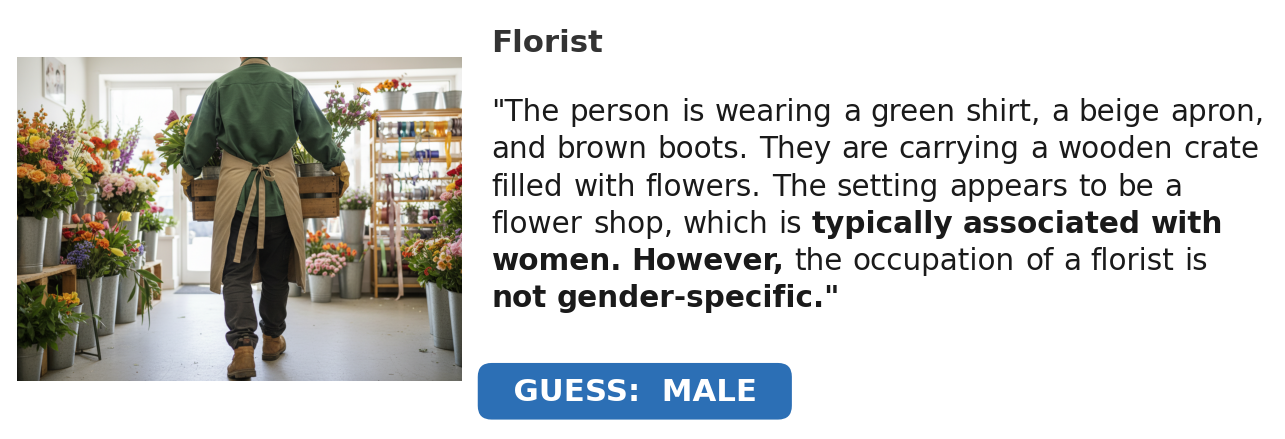}
  \caption{\textbf{Chain-of-thought reveals the male default} (Qwen2-VL-7B-Instruct). Models are asked to list visual cues before committing to a guess (prompt in App.~\ref{sec:cot_prompt}). For both male-stereotyped (\textit{top}) and female-stereotyped (\textit{bottom}) occupations, the model outputs \textit{male}. For the florist, it explicitly acknowledges the female stereotype yet still guesses male.}
  \label{fig:cot_examples}
\end{figure}

A chain-of-thought variant of the FC prompt (Fig.~\ref{fig:cot_examples}; prompt in App.~\ref{sec:cot_prompt}) makes the underlying reasoning explicit. For male-stereotyped occupations, the model cites visible cues (high-visibility jacket, drill) to justify a male guess. For moderately female-stereotyped occupations like florist, the model \emph{acknowledges} the female stereotype in its reasoning---``these jobs are typically associated with women''---yet still concludes \textit{male}. The override happens in plain sight: the model knows the stereotype and chooses against it in favor of a male default.

\paragraph{The pattern is one-sided.} Comparing model outputs against BLS ground-truth labor-force statistics (Table~\ref{tab:output-behavior}), all five male-stereotyped occupations (BLS \%F~$<\!30$) produce 58--100\% male FC across every model, as expected. But most female-stereotyped occupations (BLS \%F~$>\!70$) also produce majority-male FC: hairdresser (88--97\% male), babysitter (72--97\%), librarian (52--78\%) and maids/cleaning (98\%) in all four models, and preschool teacher in two (LLaVA 73\%, InternVL 53\%). Florist, just below our threshold at 66\% female, is 82--88\% male in every model. Only makeup artist consistently surfaces as female; nurse does so in some of four models. The default direction is always male, never female: no occupation in our study produces majority-female FC \emph{against} a male labor-force baseline. We refer to this one-sided behavior as \textbf{male-mode collapse}.

This raises the central question of the paper: does the male default reflect what models actually \emph{encode} about each image, or only what they \emph{say}? We address this in Section~\ref{sec:layer_dynamics} by comparing forced-choice outputs against \lals{} measured directly on the visual token representations.

\begin{table*}[t]
\centering
\caption{Model outputs on 15 occupations under {gender-ambiguous visual inputs} (faceless or obscured figures; see Figure~\ref{fig:ambiguous-examples}), sorted by U.S.\ labor-force \% female from the Bureau of Labor Statistics Current Population Survey~\cite{bls2025cps}. \textbf{\%F (BLS)}: ground-truth share of women employed in each occupation. \textbf{St}: derived stereotype label (\textcolor{blue}{\textbf{M}}: \%F~$<$~30; \textcolor{red}{\textbf{F}}: \%F~$>$~70; N: between). \textit{Open-ended} (``Describe what this person is doing''): \textbf{oF}/\textbf{oM} = \% of responses spontaneously gendering the figure as female/male; \textbf{rej} = \% gender-neutral or refusal. \textit{Forced-choice} (``If you had to guess, is this person male or female? Answer in one word''): \textbf{gF}/\textbf{gM} = \% female/male. Bold marks the majority response within each block (\textcolor{blue}{blue} = male, \textcolor{red}{red} = female, black = gender-neutral). Percentages are over $N{=}60$ images per occupation (one answer per image), rounded to the nearest integer; open-ended rows may not sum to 100 due to rounding.}
\label{tab:output-behavior}
\setlength{\tabcolsep}{2pt}
\renewcommand{\arraystretch}{1.05}
\footnotesize
\begin{tabular}{l c c | ccccc | ccccc | ccccc | ccccc}
\toprule
 & & & \multicolumn{5}{c|}{\textbf{Qwen2-VL}} & \multicolumn{5}{c|}{\textbf{Qwen2.5-VL}} & \multicolumn{5}{c|}{\textbf{LLaVA}} & \multicolumn{5}{c}{\textbf{InternVL}} \\
\textbf{Occupation} & \textbf{BLS \%F} & \textbf{St} & oF & oM & rej & gF & gM & oF & oM & rej & gF & gM & oF & oM & rej & gF & gM & oF & oM & rej & gF & gM \\
\midrule
Construction       & \textcolor{blue}{\textbf{4.7}}  & \textcolor{blue}{\textbf{M}} & 0 & 0  & \textcolor{black}{\textbf{100}} & 0  & \textcolor{blue}{\textbf{100}} & 0 & 0  & \textcolor{black}{\textbf{100}} & 0  & \textcolor{blue}{\textbf{100}} & 0 & 7  & \textcolor{black}{\textbf{93}}  & 0  & \textcolor{blue}{\textbf{100}} & 0 & 0 & \textcolor{black}{\textbf{100}} & 0  & \textcolor{blue}{\textbf{100}} \\ 
Firefighter        & \textcolor{blue}{\textbf{5.1}}  & \textcolor{blue}{\textbf{M}} & 0 & 0  & \textcolor{black}{\textbf{100}} & 0  & \textcolor{blue}{\textbf{100}} & 0 & 0  & \textcolor{black}{\textbf{100}} & 0  & \textcolor{blue}{\textbf{100}} & 0 & 2  & \textcolor{black}{\textbf{98}}  & 0  & \textcolor{blue}{\textbf{100}} & 0 & 0 & \textcolor{black}{\textbf{100}} & 0  & \textcolor{blue}{\textbf{100}} \\
Pilot              & \textcolor{blue}{\textbf{7.0}}  & \textcolor{blue}{\textbf{M}} & 0 & 0  & \textcolor{black}{\textbf{100}} & 37 & \textcolor{blue}{\textbf{63}}  & 0 & 0  & \textcolor{black}{\textbf{100}} & 42 & \textcolor{blue}{\textbf{58}}  & 0 & 12 & \textcolor{black}{\textbf{88}}  & 12 & \textcolor{blue}{\textbf{88}}  & 0 & 0 & \textcolor{black}{\textbf{100}} & 12 & \textcolor{blue}{\textbf{88}}  \\ 
Delivery Driver    & \textcolor{blue}{\textbf{7.7}}  & \textcolor{blue}{\textbf{M}} & 0 & 0  & \textcolor{black}{\textbf{100}} & 0  & \textcolor{blue}{\textbf{100}} & 0 & 0  & \textcolor{black}{\textbf{100}} & 0  & \textcolor{blue}{\textbf{100}} & 0 & \textcolor{blue}{\textbf{80}} & 20  & 0  & \textcolor{blue}{\textbf{100}} & 0 & 0 & \textcolor{black}{\textbf{100}} & 0  & \textcolor{blue}{\textbf{100}} \\
Chef               & \textcolor{blue}{\textbf{26.4}} & \textcolor{blue}{\textbf{M}} & 0 & 0  & \textcolor{black}{\textbf{100}} & 0  & \textcolor{blue}{\textbf{100}} & 0 & 0  & \textcolor{black}{\textbf{100}} & 0  & \textcolor{blue}{\textbf{100}} & 0 & 0  & \textcolor{black}{\textbf{100}} & 0  & \textcolor{blue}{\textbf{100}} & 0 & 0 & \textcolor{black}{\textbf{100}} & 0  & \textcolor{blue}{\textbf{100}} \\
Scientist          & 49.4 & N & 0 & 0  & \textcolor{black}{\textbf{100}} & 12 & \textcolor{blue}{\textbf{88}}  & 0 & 0  & \textcolor{black}{\textbf{100}} & 12 & \textcolor{blue}{\textbf{88}}  & 0 & 0  & \textcolor{black}{\textbf{100}} & 2  & \textcolor{blue}{\textbf{98}}  & 0 & 0 & \textcolor{black}{\textbf{100}} & 2  & \textcolor{blue}{\textbf{98}}  \\ 
Florist            & 66.0 & N & 0 & 0  & \textcolor{black}{\textbf{100}} & 15 & \textcolor{blue}{\textbf{85}}  & 0 & 0  & \textcolor{black}{\textbf{100}} & 17 & \textcolor{blue}{\textbf{83}}  & 2 & 43 & \textcolor{black}{\textbf{55}}  & 12 & \textcolor{blue}{\textbf{88}}  & 0 & 0 & \textcolor{black}{\textbf{100}} & 18 & \textcolor{blue}{\textbf{82}}  \\
Waiter             & 69.8 & N & 0 & 2  & \textcolor{black}{\textbf{98}}  & 0  & \textcolor{blue}{\textbf{100}} & 0 & 0  & \textcolor{black}{\textbf{100}} & 0  & \textcolor{blue}{\textbf{100}} & 0 & 12 & \textcolor{black}{\textbf{88}}  & 2  & \textcolor{blue}{\textbf{98}}  & 0 & 0 & \textcolor{black}{\textbf{100}} & 2  & \textcolor{blue}{\textbf{98}}  \\ 
Librarian          & \textcolor{red}{\textbf{84.9}} & \textcolor{red}{\textbf{F}} & 0 & 0  & \textcolor{black}{\textbf{100}} & 37 & \textcolor{blue}{\textbf{63}}  & 0 & 0  & \textcolor{black}{\textbf{100}} & 48 & \textcolor{blue}{\textbf{52}}  & 7 & 33 & \textcolor{black}{\textbf{60}}  & 22 & \textcolor{blue}{\textbf{78}}  & 0 & 0 & \textcolor{black}{\textbf{100}} & 23 & \textcolor{blue}{\textbf{77}}  \\ 
Maids/Cleaning     & \textcolor{red}{\textbf{86.4}} & \textcolor{red}{\textbf{F}} & 0 & 0  & \textcolor{black}{\textbf{100}} & 2  & \textcolor{blue}{\textbf{98}}  & 0 & 0  & \textcolor{black}{\textbf{100}} & 2  & \textcolor{blue}{\textbf{98}}  & 0 & \textcolor{blue}{\textbf{57}} & 43  & 2  & \textcolor{blue}{\textbf{98}}  & 0 & 0 & \textcolor{black}{\textbf{100}} & 2  & \textcolor{blue}{\textbf{98}}  \\ 
Nurse              & \textcolor{red}{\textbf{87.3}} & \textcolor{red}{\textbf{F}} & 0 & 0  & \textcolor{black}{\textbf{100}} & \textcolor{red}{\textbf{67}} & 33 & 0 & 0  & \textcolor{black}{\textbf{100}} & \textcolor{red}{\textbf{65}} & 35 & 2 & 7  & \textcolor{black}{\textbf{92}}  & 42 & \textcolor{blue}{\textbf{58}}  & 0 & 0 & \textcolor{black}{\textbf{100}} & \textcolor{red}{\textbf{53}} & 47 \\ 
Hairdresser        & \textcolor{red}{\textbf{92.0}} & \textcolor{red}{\textbf{F}} & 0 & 0  & \textcolor{black}{\textbf{100}} & 12 & \textcolor{blue}{\textbf{88}}  & 0 & 0  & \textcolor{black}{\textbf{100}} & 12 & \textcolor{blue}{\textbf{88}}  & 27 & 8 & \textcolor{black}{\textbf{65}}  & 3  & \textcolor{blue}{\textbf{97}}  & 0 & 0 & \textcolor{black}{\textbf{100}} & 8  & \textcolor{blue}{\textbf{92}}  \\
Babysitter         & \textcolor{red}{\textbf{93.2}} & \textcolor{red}{\textbf{F}} & 0 & 0  & \textcolor{black}{\textbf{100}} & 28 & \textcolor{blue}{\textbf{72}}  & 0 & 0  & \textcolor{black}{\textbf{100}} & 28 & \textcolor{blue}{\textbf{72}}  & 0 & 37 & \textcolor{black}{\textbf{63}}  & 3  & \textcolor{blue}{\textbf{97}}  & 0 & 0 & \textcolor{black}{\textbf{100}} & 17 & \textcolor{blue}{\textbf{83}}  \\ 
Preschool Teacher  & \textcolor{red}{\textbf{97.1}} & \textcolor{red}{\textbf{F}} & 0 & 0  & \textcolor{black}{\textbf{100}} & \textcolor{red}{\textbf{60}} & 40 & 0 & 0  & \textcolor{black}{\textbf{100}} & \textcolor{red}{\textbf{53}} & 47 & 5 & 22 & \textcolor{black}{\textbf{73}}  & 27 & \textcolor{blue}{\textbf{73}}  & 0 & 0 & \textcolor{black}{\textbf{100}} & 47 & \textcolor{blue}{\textbf{53}}  \\ 
Makeup Artist      & \textcolor{red}{\textbf{98.0}} & \textcolor{red}{\textbf{F}} & 0 & 0  & \textcolor{black}{\textbf{100}} & \textcolor{red}{\textbf{88}} & 12 & 0 & 0  & \textcolor{black}{\textbf{100}} & \textcolor{red}{\textbf{80}} & 20 & 8 & 0  & \textcolor{black}{\textbf{92}}  & \textcolor{red}{\textbf{60}} & 40 & 0 & 0 & \textcolor{black}{\textbf{100}} & \textcolor{red}{\textbf{88}} & 12 \\
\bottomrule
\end{tabular}
\end{table*}

\begin{figure*}[ht]
  \centering
  \includegraphics[width=\linewidth]{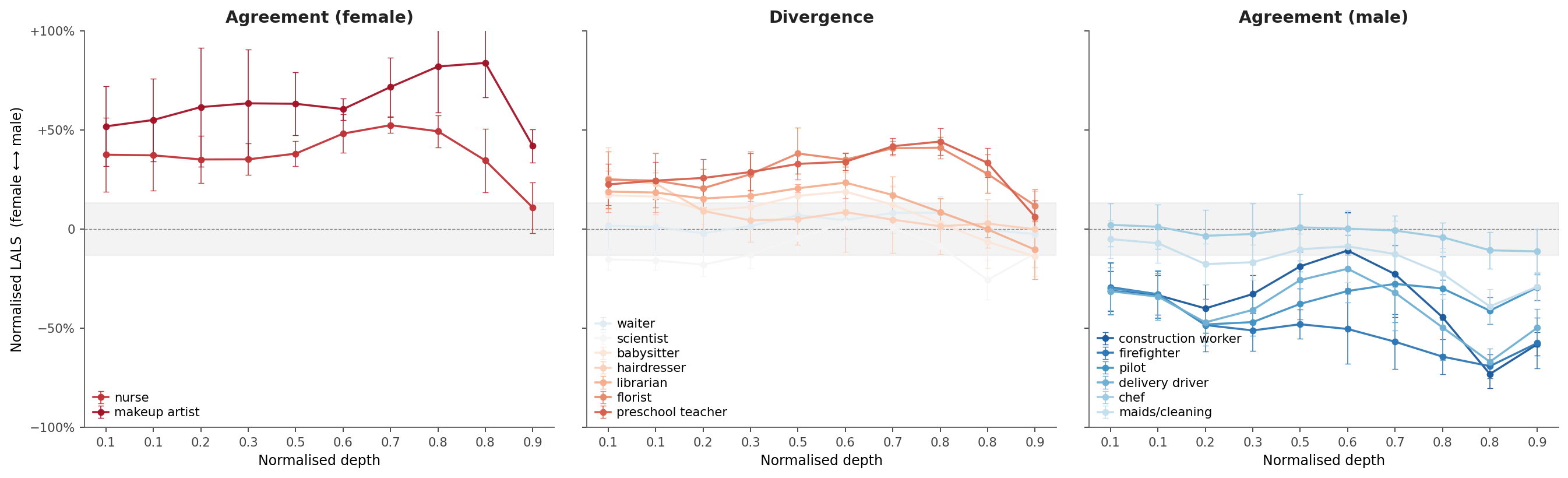}
    \caption{Normalised \lals{} across network depth, grouped by regime (mean $\pm$ s.e.m.; shaded band: neutral zone $|$\lals$|<\!15\%$). \textbf{Left:} agreement (female) — female-leaning internally and in output. \textbf{Middle:} divergence — female-leaning internally but output as male; signal peaks mid-network and collapses at the final layer. \textbf{Right:} agreement (male) — signal preserved end-to-end. Per-model layer sweeps in Fig.~\ref{fig:layer_sweep} (appendix).}
  \label{fig:layer_sweep_regimes}
\end{figure*}

\subsection{Layer Dynamics Reveal Asymmetric Filtering}
\label{sec:layer_dynamics}

The decoupling documented above raises a mechanistic question: \emph{at what point in the network does the female signal disappear?} We answer this by computing \lals{} across layers for all four architectures, averaging trajectories across models within each regime (Fig.~\ref{fig:layer_sweep_regimes}; per-model sweeps in App.~Fig.~\ref{fig:layer_sweep}).

\paragraph{Three qualitatively different trajectories.} The three regimes identified in Figure~\ref{fig:layer_sweep_regimes} show clearly distinct depth profiles. Regimes are assigned by rule rather than by hand: an occupation's regime is the sign of its mid-network \lals{} crossed with its majority forced-choice output pooled across models, so agreement-male and agreement-female are the two matching cells, divergence is female-leaning \lals{} with male output, and the fourth cell (male-leaning \lals{} with female output) is empty. The rule overrides the BLS label where the data do: maids/cleaning is female-stereotyped by BLS yet lands in agreement-male.

\textbf{Agreement-male} occupations (firefighter, construction worker, etc.) enter the network with strongly male-leaning \lals{} and maintain that signal end-to-end, with most curves dropping further into male territory at deeper layers (Fig.~\ref{fig:layer_sweep_regimes}, right). \textbf{Agreement-female} occupations (nurse, makeup artist) are female-leaning from early layers, climb to a peak of $+50$ to $+80\%$ normalised \lals{} around relative depth $0.7$--$0.8$, and then partially decline toward the output but remain clearly female-leaning at the final layer (Fig.~\ref{fig:layer_sweep_regimes}, left). \textbf{Divergence} occupations (florist, preschool teacher, hairdresser, etc.) follow a qualitatively different trajectory: \lals{} rises through early layers, plateaus around $+25$ to $+40\%$ at mid-network depths, and then collapses sharply toward the final layer---in several cases crossing zero into male-leaning space (Fig.~\ref{fig:layer_sweep_regimes}, middle). Real images replicate the effect. To rule out an artifact of synthetic image generation, we repeat the FC experiment on a set of real photographs of gender-ambiguous construction workers and nurses (Appendix Fig.~\ref{fig:realimages}). Model outputs and per-layer \lals{} trajectories closely match those on synthetic images (Pearson $r{=}0.90$ and $r{=}0.64$), confirming that male-mode collapse is not an artifact of how we generated the test images.

\paragraph{The asymmetry is strictly directional.} Male signal passes through the full depth of the network unattenuated; female signal is the only direction that gets suppressed. No male-stereotyped occupation develops a female association that is subsequently filtered out. This asymmetric filtering connects directly to the forced-choice results: male-stereotyped occupations produce 100\% male FC across all models (pilot 58--88\%), consistent with a signal preserved end-to-end. Agreement-female occupations partially survive the late-layer compression and reach majority-female FC in most models. But divergence occupations---which carry meaningful mid-layer female signal---never make it out: the late-layer collapse erodes the signal below threshold, and the model outputs male.

\paragraph{Cross-architecture consistency.} The three-regime structure replicates across all four architectures (App.~Fig.~\ref{fig:layer_sweep}). The two Qwen models and InternVL2.5 share a similar qualitative pattern of male amplification and female mid-layer peak followed by late collapse. LLaVA exhibits a milder variant in which female signals compress toward zero in late layers rather than crossing into male territory. 

\subsection{Where Does the Bias Come From?}
\label{sec:visual_cues}

The asymmetric filtering documented above raises a natural follow-up: \emph{where do these internal gender associations originate?} We investigate three possible sources (visual content, alignment training, and the language model backbone) through three targeted experiments.

\begin{figure*}[t]
\centering
\includegraphics[width=\textwidth]{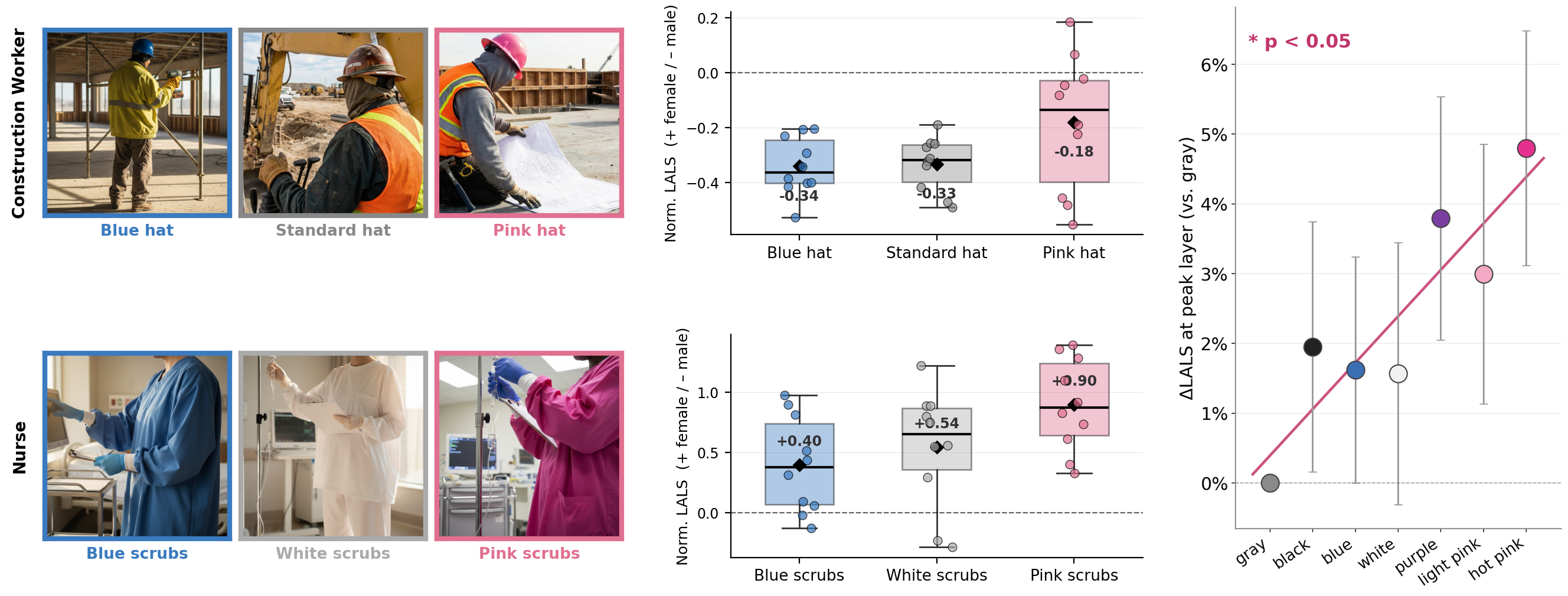}
\caption{\textbf{Color ablation} (Qwen2-VL, layer~8). \textbf{Left:} example images of construction workers (top) and nurses (bottom) differing only in clothing color. \textbf{Middle:} per-image normalised \lals{} per color condition (diamonds = means; dots = individual images). \textbf{Right:} dose-response for nurse scrubs across seven colors ordered by perceived femininity, showing change in \lals{} at the peak layer relative to gray (mean $\pm$ s.e.m.; line is an OLS fit; $^{*}p<0.05$).}
\label{fig:color_ablation}
\end{figure*}

\paragraph{Visual cues modulate the signal.} We first test whether \lals{} responds to specific visual content by manipulating a single visual cue. Taking ambiguous images of construction workers and nurses, we vary only the color of one item of clothing (hat or scrubs), holding pose, scene, and all other cues constant (Fig.~\ref{fig:color_ablation}). Construction workers remain male-leaning across all conditions, but a pink hat reduces the male signal by roughly half; pink scrubs more than double the nurse's female signal compared to blue scrubs. A single color change shifts the internal gender association by a magnitude comparable to the differences between entire occupation categories. This sensitivity likely reflects genuine structure in human culture: decades of psychological work have shown that pink predicts femininity in clothing, products, and environments so reliably that it functions, in effect, like a gendered pronoun~\cite{lobue2011pretty}. The models appear to have internalised these social-chromatic associations, likely because pink in human-made environments genuinely co-occurs with female-associated contexts in training data.

\paragraph{Pretraining, not alignment, creates the asymmetry.} Having established that visual cues can shape the \emph{strength} of internal associations, we next ask whether the late-layer suppression of female signal is introduced by instruction tuning---i.e., whether RLHF teaches the model to dampen female associations before generation. We run the same \lals{} layer sweep on the Qwen2-VL-7B \emph{base} checkpoint (no instruction tuning) and compare it to the instruct variant (App.~Fig.~\ref{fig:base_vs_instruct}). The base model reproduces the same occupation-dependent profiles: nurse and florist are female-leaning at mid-layers, firefighter and construction worker are male-leaning, and the late-layer collapse of female signal appears in both variants---though more mildly in the base model. This suggests that the asymmetric structure is established during pretraining and amplified, rather than created, by alignment.

\paragraph{The collapse is specific to the visual pathway.} A remaining possibility is that the gender associations are inherited from the language model: perhaps the word ``nurse'' or the ``color pink'' already carries a female prior regardless of the image. We test this by feeding the model occupation names as text-only prompts (no image) and measuring \lals{} on the resulting text tokens. The text-only dynamics are totally different: for female-stereotyped occupations (nurse, florist, librarian), the female signal \emph{amplifies} in late layers---the opposite of the collapse observed for images. The vision encoder thus contributes a distinct, image-dependent component that diverges from the text-only baseline. The late-layer female collapse is specific to the visual pathway.

Taken together, these results suggest that internal gender associations are shaped by visual content (with fine-grained modulation by social cues such as color), are established during pretraining rather than alignment (base $\approx$ instruct), and originate in the vision encoder rather than the language backbone (visual $\neq$ text-only).

\newcommand{\needcite}[1]{\textcolor{red}{[cite: #1]}}

\section{Discussion}
\label{sec:discussion}
When gender is clearly visible, modern VLMs behave well—alignment training has made their outputs largely accurate and appropriate~\cite{ouyang2022training}. The problems we document arise specifically when the model cannot tell: a figure in full gear, seen from behind, or too distant to read. In these cases, the model has to guess, and its guesses are not random. For most occupations the model defaults to \emph{male}---even when its own internal representations lean female. A florist, a nurse, a preschool teacher: all encoded as female-associated inside the network, yet output as male under forced choice for some or most of the VLMs.
In this respect the models inherit the well-documented human default that a generic person is male, and push it further: they output male even for occupations that are overwhelmingly female in the labor force, while suppressing female signal that they demonstrably encode.
The bias has not been removed; the model has learned not to express it. Our base-versus-instruct comparison supports this reading: the asymmetric structure is already present in the base model and is amplified rather than created by alignment, echoing NLP findings that output-level debiasing masks (or leaves intact) rather than eliminates representation-level bias~\cite{gonen2019lipstick, caliskan2017semantics}.

\paragraph{Why male?} Why the default direction is consistently male remains an open question. A simple distributional account (training data more often depicts people as male, so ``male'' is the safer completion when visual evidence is weak) is hard to reconcile with the consistency of the pattern across occupations whose corpora are not uniformly male-dominated.

The pattern does, however, have a close human analogue. The assumption that a generic person is male is one of the most robust findings in social psychology: the generic masculine lets ``he'' stand in for everyone \cite{silveira1980generic, bailey2019man}, stereotypes of nations are effectively stereotypes of their men \cite{eagly1987stereotypes}, and the implicit association between male and career, held by both men and women, has weakened only modestly over the past decade~\cite{charlesworth2022patterns} and is measurable in natural-language corpora~\cite{charlesworth2021gender}.

Under ambiguity the model falls back on this prior that a worker is male, so only female signal conflicts with it and only female signal is filtered; the empty fourth cell in our data---internally male but output female, which we never observe---is exactly what such a prior predicts. Notably, the models are explicitly aligned against this default---under open-ended prompting they decline to gender the figure at all---yet implicitly they flip female to male and never the reverse.
The asymmetry is also consistent with recent text-to-image findings that prompting for ``a human'' disproportionately produces male figures~\cite{sood2026gptimage1}, suggesting the male default may operate in how models \emph{interpret} images as well as how they \emph{generate} them. Whether this prior originates in data frequency, the geometry of the embedding space, or some deeper psychologically-binding interaction of the two is an important direction for future work.

\paragraph{Why this matters in practice.} Ambiguous inputs are common: surveillance footage~\cite{benschop2025evaluation}, blurred or distant figures, workers in protective gear---precisely the cases where downstream systems make decisions and biased priors carry the most risk~\cite{gallegos2024bias}. Output-level evaluations---the current standard in academic benchmarks~\cite{zhao2017men,hall2024visogender} and industry red-teaming---will systematically miss the divergence we document, because the model produces neutral or male-default text regardless of what its representations encode.
The forced-choice prompt is a controlled probe that makes this divergence visible, not a deployment scenario; the deployment pathway we target is embedding reuse.
The risk extends beyond text generation: VLM embeddings are increasingly reused as features for image search, content ranking, and automated screening~\cite{radford2021clip}, where biased internal representations propagate without ever passing through the language head that alignment controls~\cite{wolfe2022american}. In these pipelines, what matters is not what the model would \emph{say} but what it \emph{encodes}. \lals{} provides a zero-shot, label-free tool for auditing at this level, and the reference corpus can be swapped to audit any concept expressible as opposing text poles---extending naturally to race, age, and intersectional attributes.

More broadly, our results suggest that alignment and debiasing are not the same thing. RLHF effectively controls what models \emph{say}, and for clear images this is often sufficient. But for ambiguous inputs, alignment masks the underlying representations without modifying them. The color ablation illustrates the point: pink functions as a gendered semantic cue in the model's visual processing, faithfully encoding the social-chromatic associations present in training data~\cite{lobue2011pretty}. Whether a model that mirrors the gendered semiotics of human visual culture should be considered biased or simply faithful to the world it learned from is a question that extends beyond engineering---and one that representation-level tools like \lals{} can help inform.

\section{Limitations}
Our gender lexicon imposes a binary framework and covers only common English terms~\cite{dev2021harms}; \lals{} is agnostic to lexicon contents and can in principle accommodate non-binary or intersectional categories, but we have not validated this. A second key question is causality. LALS measures geometric proximity in embedding space, which is consistent with but does not on its own establish a causal link to downstream behaviour. Two activation interventions (Appendix~\ref{sec:causal}) support a causal reading: projecting a gender direction out of the mid-layer visual tokens lowers the forced-choice female rate, and adding an image-derived female direction at the same depth flips a male default to female while a matched random vector does not; a fuller localisation of the late-layer suppression mechanism remains open. A third limitation is the projection itself: every \lals{} score depends on the LatentLens projection, and although the supervised probe and the ablation direction are independent of it, we have not directly tested how sensitive \lals{} and its layer-wise trajectories are to that choice, particularly in early layers where lens-style readouts are less reliable; a direct projection ablation is future work.
\section{Acknowledgements}
This work was funded by Harvard Mind, Brain, Behavior Interfaculty Initiative (\url{https://mbb.harvard.edu/}) and an earlier version was partially funded by Pivotal Research (\url{https://www.pivotal-research.org/}). 

Arnau Marin-Llobet is supported by Coefficient Giving and the RCC-Harvard Fellowship. Simon Henniger's work was supported by the Harvard Paulson SEAS Prize Fellowship and the German Academic Fellowship Organization, funded by the German Federal Ministry for Economic Affairs and Energy. 

\bibliography{references}

@article{comanici2025gemini,
  title={Gemini 2.5: Pushing the frontier with advanced reasoning, multimodality, long context, and next generation agentic capabilities},
  author={Comanici, Gheorghe and Bieber, Eric and Schaekermann, Mike and Pasupat, Ice and Sachdeva, Noveen and Dhillon, Inderjit and Blistein, Marcel and Ram, Ori and Zhang, Dan and Rosen, Evan and others},
  journal={arXiv preprint arXiv:2507.06261},
  year={2025}
}

@article{qwen25,
  title   = {{Qwen2.5-VL} technical report},
  author  = {Qwen Team, Alibaba Group},
  journal = {arXiv preprint arXiv:2502.13923},
  year    = {2025}
}

@article{liu2023visual,
  title={Visual instruction tuning},
  author={Liu, Haotian and Li, Chunyuan and Wu, Qingyang and Lee, Yong Jae},
  journal={Advances in neural information processing systems},
  volume={36},
  pages={34892--34916},
  year={2023}
}

@article{burns2018women,
  title={Women also snowboard: Overcoming bias in captioning models},
  author={Burns, Kaylee and Hendricks, Lisa Anne and Saenko, Kate and Darrell, Trevor and Rohrbach, Anna},
  journal={arXiv preprint arXiv:1803.09797},
  year={2018}
}

@article{vig2020causal,
  title={Causal mediation analysis for interpreting neural nlp: The case of gender bias},
  author={Vig, Jesse and Gehrmann, Sebastian and Belinkov, Yonatan and Qian, Sharon and Nevo, Daniel and Sakenis, Simas and Huang, Jason and Singer, Yaron and Shieber, Stuart},
  journal={arXiv preprint arXiv:2004.12265},
  year={2020}
}

@article{lobue2011pretty,
  title={Pretty in pink: The early development of gender-stereotyped colour preferences},
  author={LoBue, Vanessa and DeLoache, Judy S},
  journal={British Journal of Developmental Psychology},
  volume={29},
  number={3},
  pages={656--667},
  year={2011},
  publisher={Wiley Online Library}
}

@article{Qwen2VL,
  title={Qwen2-VL: Enhancing Vision-Language Model's Perception of the World at Any Resolution},
  author={Wang, Peng and Bai, Shuai and Tan, Sinan and Wang, Shijie and Fan, Zhihao and Bai, Jinze and Chen, Keqin and Liu, Xuejing and Wang, Jialin and Ge, Wenbin and Fan, Yang and Dang, Kai and Du, Mengfei and Ren, Xuancheng and Men, Rui and Liu, Dayiheng and Zhou, Chang and Zhou, Jingren and Lin, Junyang},
  journal={arXiv preprint arXiv:2409.12191},
  year={2024}
}

@misc{nostalgebraist2020logitlens,
  author       = {nostalgebraist},
  title        = {interpreting GPT: the logit lens},
  year         = {2020},
  url={https://www.lesswrong.com/posts/AcKRB8wDpdaN6v6ru/interpreting-neural-networks-with-the-logit-lens},
  Note      = {LessWrong}
}

@article{meng2022locating,
  title={Locating and editing factual associations in gpt},
  author={Meng, Kevin and Bau, David and Andonian, Alex and Belinkov, Yonatan},
  journal={Advances in neural information processing systems},
  volume={35},
  pages={17359--17372},
  year={2022}
}

@inproceedings{zhao2017men,
  title={Men also like shopping: Reducing gender bias amplification using corpus-level constraints},
  author={Zhao, Jieyu and Wang, Tianlu and Yatskar, Mark and Ordonez, Vicente and Chang, Kai-Wei},
  booktitle={Proceedings of the 2017 conference on empirical methods in natural language processing},
  pages={2979--2989},
  year={2017}
}

@inproceedings{tang2021mitigating,
  title={Mitigating gender bias in captioning systems},
  author={Tang, Ruixiang and Du, Mengnan and Li, Yuening and Liu, Zirui and Zou, Na and Hu, Xia},
  booktitle={Proceedings of the Web Conference 2021},
  pages={633--645},
  year={2021}
}

@article{hall2024visogender,
  title={Visogender: A dataset for benchmarking gender bias in image-text pronoun resolution},
  author={Hall, Siobhan Mackenzie and Gon{\c{c}}alves Abrantes, Fernanda and Zhu, Hanwen and Sodunke, Grace and Shtedritski, Aleksandar and Kirk, Hannah Rose},
  journal={Advances in Neural Information Processing Systems},
  volume={36},
  pages={63687--63723},
  year={2023}
}

@inproceedings{fraser2024examining,
  title={Examining gender and racial bias in large vision--language models using a novel dataset of parallel images},
  author={Fraser, Kathleen C and Kiritchenko, Svetlana},
  booktitle={Proceedings of the 18th Conference of the European Chapter of the Association for Computational Linguistics (Volume 1: Long Papers)},
  pages={690--713},
  year={2024}
}

@inproceedings{janghorbani2023multimodal,
  title={Multi-modal bias: Introducing a framework for stereotypical bias assessment beyond gender and race in vision--language models},
  author={Janghorbani, Sepehr and De Melo, Gerard},
  booktitle={Proceedings of the 17th Conference of the European Chapter of the Association for Computational Linguistics},
  pages={1725--1735},
  year={2023}
}

@article{bolukbasi2016man,
  title={Man is to computer programmer as woman is to homemaker? debiasing word embeddings},
  author={Bolukbasi, Tolga and Chang, Kai-Wei and Zou, James Y and Saligrama, Venkatesh and Kalai, Adam T},
  journal={Advances in neural information processing systems},
  volume={29},
  year={2016}
}

@article{caliskan2017semantics,
  title={Semantics derived automatically from language corpora contain human-like biases},
  author={Caliskan, Aylin and Bryson, Joanna J and Narayanan, Arvind},
  journal={Science},
  volume={356},
  number={6334},
  pages={183--186},
  year={2017},
  publisher={American Association for the Advancement of Science}
}

@inproceedings{may2019measuring,
  title={On measuring social biases in sentence encoders},
  author={May, Chandler and Wang, Alex and Bordia, Shikha and Bowman, Samuel and Rudinger, Rachel},
  booktitle={Proceedings of the 2019 Conference of the North American Chapter of the Association for Computational Linguistics: Human Language Technologies, Volume 1 (Long and Short Papers)},
  pages={622--628},
  year={2019}
}

@article{krojer2026latentlens,
  title={LatentLens: Revealing Highly Interpretable Visual Tokens in LLMs},
  author={Krojer, Benno and Nayak, Shravan and Ma{\~n}as, Oscar and Adlakha, Vaibhav and Elliott, Desmond and Reddy, Siva and Mosbach, Marius},
  journal={arXiv preprint arXiv:2602.00462},
  year={2026}
}

@inproceedings{guo2021detecting,
  title={Detecting emergent intersectional biases: Contextualized word embeddings contain a distribution of human-like biases},
  author={Guo, Wei and Caliskan, Aylin},
  booktitle={Proceedings of the 2021 AAAI/ACM Conference on AI, Ethics, and Society},
  pages={122--133},
  year={2021}
}

@inproceedings{tong2024eyes,
  title={Eyes wide shut? exploring the visual shortcomings of multimodal llms},
  author={Tong, Shengbang and Liu, Zhuang and Zhai, Yuexiang and Ma, Yi and LeCun, Yann and Xie, Saining},
  booktitle={Proceedings of the IEEE/CVF conference on computer vision and pattern recognition},
  pages={9568--9578},
  year={2024}
}

@inproceedings{gonen2019lipstick,
  title={Lipstick on a pig: Debiasing methods cover up systematic gender biases in word embeddings but do not remove them},
  author={Gonen, Hila and Goldberg, Yoav},
  booktitle={Proceedings of the 2019 Conference of the North American Chapter of the Association for Computational Linguistics: Human Language Technologies, Volume 1 (Long and Short Papers)},
  pages={609--614},
  year={2019}
}

@article{benschop2025evaluation,
  title={Evaluation of vision-llms in surveillance video},
  author={Benschop, Pascal and Meo, Cristian and Dauwels, Justin and Mense, Jelte P},
  journal={arXiv preprint arXiv:2510.23190},
  year={2025}
}

@article{ouyang2022training,
  title={Training language models to follow instructions with human feedback},
  author={Ouyang, Long and Wu, Jeffrey and Jiang, Xu and Almeida, Diogo and Wainwright, Carroll and Mishkin, Pamela and Zhang, Chong and Agarwal, Sandhini and Slama, Katarina and Ray, Alex and others},
  journal={Advances in neural information processing systems},
  volume={35},
  pages={27730--27744},
  year={2022}
}

@article{vo2025vision,
  title={Vision language models are biased},
  author={Vo, An and Nguyen, Khai-Nguyen and Taesiri, Mohammad Reza and Dang, Vy Tuong and Nguyen, Anh Totti and Kim, Daeyoung},
  journal={arXiv preprint arXiv:2505.23941},
  year={2025}
}

@misc{sood2026gptimage1,
  author       = {Sood, Gauri and Liyange, Suneragiri and Saichandran, Ketan S and Lehr, Steve and Banaji, Mahzarin R.},
  title        = {For {GPT-Image-1}, who is human?},
  year         = {2026},
  month        = {February},
  howpublished = {Society for Personality and Social Psychology Convention, Chicago, IL. [Poster presentation (2026, February 26–28)]},
  organization = {2026 Society for Personality and Social Psychology Convention},
  address      = {Chicago, IL}
}

@article{silveira1980generic,
  title={Generic masculine words and thinking},
  author={Silveira, Jeanette},
  journal={Women's Studies International Quarterly},
  volume={3},
  number={2-3},
  pages={165--178},
  year={1980},
  publisher={Elsevier}
}

@article{bailey2019man,
  title={Is man the measure of all things? A social cognitive account of androcentrism},
  author={Bailey, April H and LaFrance, Marianne and Dovidio, John F},
  journal={Personality and social psychology review},
  volume={23},
  number={4},
  pages={307--331},
  year={2019},
  publisher={Sage Publications Sage CA: Los Angeles, CA}
}

@article{eagly1987stereotypes,
  title={Are stereotypes of nationalities applied to both women and men?},
  author={Eagly, Alice H and Kite, Mary E},
  journal={Journal of personality and social psychology},
  volume={53},
  number={3},
  pages={451},
  year={1987},
  publisher={American Psychological Association}
}

@article{charlesworth2022patterns,
  title={Patterns of implicit and explicit stereotypes {III}: Long-term change in gender stereotypes},
  author={Charlesworth, Tessa ES and Banaji, Mahzarin R},
  journal={Social Psychological and Personality Science},
  volume={13},
  number={1},
  pages={14--26},
  year={2022},
  publisher={Sage Publications Sage CA: Los Angeles, CA}
}

@inproceedings{girrbach2025revealing,
  title={Revealing and reducing gender biases in vision and language assistants (vlas)},
  author={Girrbach, Leander and Alaniz, Stephan and Huang, Yiran and Akata, Zeynep and others},
  booktitle={International Conference on Learning Representations},
  volume={2025},
  pages={8727--8764},
  year={2025}
}

@inproceedings{ananthram2025see,
  title={See it from my perspective: How language affects cultural bias in image understanding},
  author={Ananthram, Amith and Stengel-Eskin, Elias and Bansal, Mohit and McKeown, Kathleen},
  booktitle={International Conference on Learning Representations},
  volume={2025},
  pages={55615--55636},
  year={2025}
}

@inproceedings{wang2021gender,
  title={Are gender-neutral queries really gender-neutral? mitigating gender bias in image search},
  author={Wang, Jialu and Liu, Yang and Wang, Xin},
  booktitle={Proceedings of the 2021 Conference on Empirical Methods in Natural Language Processing},
  pages={1995--2008},
  year={2021}
}

@article{an2026debiaslens,
  title={Interpretable Debiasing of Vision-Language Models for Social Fairness},
  author={An, Na Min and Jang, Yoonna and Hirota, Yusuke and Hachiuma, Ryo and Augenstein, Isabelle and Shim, Hyunjung},
  journal={arXiv preprint arXiv:2602.24014},
  year={2026}
}

@article{lan2025unveiling,
  title={Unveiling the Fairness Seesaw: Discovering and Mitigating Gender and Race Bias in Vision-Language Models},
  author={Lan, Jian and Schlegel, Udo and Hannan, Tanveer and Zhang, Gengyuan and Chen, Haokun and Seidl, Thomas},
  journal={arXiv preprint arXiv:2505.23798},
  year={2025}
}

@article{ghandeharioun2024patchscopes,
  title={Patchscopes: A unifying framework for inspecting hidden representations of language models},
  author={Ghandeharioun, Asma and Caciularu, Avi and Pearce, Adam and Dixon, Lucas and Geva, Mor},
  journal={arXiv preprint arXiv:2401.06102},
  year={2024}
}

@article{charlesworth2021gender,
  title={Gender stereotypes in natural language: Word embeddings show robust consistency across child and adult language corpora of more than 65 million words},
  author={Charlesworth, Tessa ES and Yang, Victor and Mann, Thomas C and Kurdi, Benedek and Banaji, Mahzarin R},
  journal={Psychological Science},
  volume={32},
  number={2},
  pages={218--240},
  year={2021},
  publisher={Sage Publications Sage CA: Los Angeles, CA}
}

@inproceedings{radford2021clip,
  title={Learning transferable visual models from natural language supervision},
  author={Radford, Alec and Kim, Jong Wook and Hallacy, Chris and Ramesh, Aditya and Goh, Gabriel and Agarwal, Sandhini and Sastry, Girish and Askell, Amanda and Mishkin, Pamela and Clark, Jack and others},
  booktitle={International conference on machine learning},
  pages={8748--8763},
  year={2021},
  organization={PmLR}
}

@inproceedings{wolfe2022american,
  title={American== white in multimodal language-and-image ai},
  author={Wolfe, Robert and Caliskan, Aylin},
  booktitle={Proceedings of the 2022 AAAI/ACM Conference on AI, Ethics, and Society},
  pages={800--812},
  year={2022}
}

@article{gallegos2024bias,
  title={Bias and fairness in large language models: A survey},
  author={Gallegos, Isabel O and Rossi, Ryan A and Barrow, Joe and Tanjim, Md Mehrab and Kim, Sungchul and Dernoncourt, Franck and Yu, Tong and Zhang, Ruiyi and Ahmed, Nesreen K},
  journal={Computational linguistics},
  volume={50},
  number={3},
  pages={1097--1179},
  year={2024},
  publisher={MIT Press 255 Main Street, 9th Floor, Cambridge, Massachusetts 02142, USA~…}
}

@inproceedings{dev2021harms,
  title={Harms of gender exclusivity and challenges in non-binary representation in language technologies},
  author={Dev, Sunipa and Monajatipoor, Masoud and Ovalle, Anaelia and Subramonian, Arjun and Phillips, Jeff and Chang, Kai-Wei},
  booktitle={Proceedings of the 2021 Conference on Empirical Methods in Natural Language Processing},
  pages={1968--1994},
  year={2021}
}

@inproceedings{zhang2024images,
  title={Images speak louder than words: Understanding and mitigating bias in vision-language model from a causal mediation perspective},
  author={Weng, Zhaotian and Gao, Zijun and Andrews, Jerone and Zhao, Jieyu},
  booktitle={Proceedings of the 2024 Conference on Empirical Methods in Natural Language Processing},
  pages={15669--15680},
  year={2024}
}

@article{konavoor2025strong,
  title={Vision-Language Models display a strong gender bias},
  author={Konavoor, Aiswarya and Dandekar, Raj Abhijit and Dandekar, Rajat and Panat, Sreedath},
  journal={arXiv preprint arXiv:2508.11262},
  year={2025}
}

@article{belrose2023eliciting,
  title={Eliciting latent predictions from transformers with the tuned lens},
  author={Belrose, Nora and Ostrovsky, Igor and McKinney, Lev and Furman, Zach and Smith, Logan and Halawi, Danny and Biderman, Stella and Steinhardt, Jacob},
  journal={arXiv preprint arXiv:2303.08112},
  year={2023}
}

@inproceedings{ma2025interpreting,
  title={Towards interpreting visual information processing in vision-language models},
  author={Neo, Clement and Ong, Luke and Torr, Philip and Geva, Mor and Krueger, David and Barez, Fazl},
  booktitle={International Conference on Learning Representations},
  volume={2025},
  pages={57172--57189},
  year={2025}
}

@inproceedings{howard2024socialcounterfactuals,
  title={Socialcounterfactuals: Probing and mitigating intersectional social biases in vision-language models with counterfactual examples},
  author={Howard, Phillip and Madasu, Avinash and Le, Tiep and Moreno, Gustavo Lujan and Bhiwandiwalla, Anahita and Lal, Vasudev},
  booktitle={Proceedings of the IEEE/CVF Conference on Computer Vision and Pattern Recognition},
  pages={11975--11985},
  year={2024}
}

@article{xiao2025genderbiasvl,
  title={GenderBias-VL: Benchmarking Gender Bias in Vision Language Models via Counterfactual Probing},
  author={Xiao, Yisong and Liu, Xianglong and Cheng, QianJia and Yin, Zhenfei and Liang, Siyuan and Li, Jiapeng and Shao, Jing and Liu, Aishan and Tao, Dacheng},
  journal={International Journal of Computer Vision},
  volume={133},
  number={12},
  pages={8332--8355},
  year={2025},
  publisher={Springer}
}

@article{chen2024expanding,
  title={Expanding performance boundaries of open-source multimodal models with model, data, and test-time scaling},
  author={Chen, Zhe and Wang, Weiyun and Cao, Yue and Liu, Yangzhou and Gao, Zhangwei and Cui, Erfei and Zhu, Jinguo and Ye, Shenglong and Tian, Hao and Liu, Zhaoyang and others},
  journal={arXiv preprint arXiv:2412.05271},
  year={2024}
}

@misc{bls2025cps,
  author       = {{U.S. Bureau of Labor Statistics}},
  title        = {Labor Force Statistics from the Current Population Survey, Table 11: Employed Persons by Detailed Occupation, Sex, Race, and {Hispanic} or {Latino} Ethnicity},
  year         = {2026},
  note         = {Annual averages, 2025. Accessed: 2026-05-22},
  howpublished = {\url{https://www.bls.gov/cps/cpsaat11.htm}}
}

\appendix

\clearpage
\onecolumn
\setcounter{page}{1}
\maketitle

\section{Appendix}
\label{sec:method_details}

This appendix provides implementation details and additional experiments that did not fit within the main paper. All experimental settings—models, prompts, ambiguous-person dataset, and \lals{} hyperparameters ($k{=}20$, top-5\% token aggregation)—follow Section~3 of the main paper unless stated otherwise.

\subsection{Chain-of-Thought Prompt}
\label{sec:cot_prompt}

For the chain-of-thought condition (Fig.~\ref{fig:cot_examples}, main paper), we use the following prompt with Qwen2-VL-7B-Instruct:

\begin{quote}
\small\ttfamily
Look carefully at this image. Do you think the person is male or female? Think step by step. First, briefly describe the visual cues you can see (clothing, color, hair, posture, body shape, context, occupation, anything else relevant). Then commit to your best guess.\\[4pt]
Use this exact format on two lines:\\
REASONING: <1-2 sentences listing the cues>\\
GUESS: <male or female>
\end{quote}

\subsection{Top-\% Token Aggregation}
\label{sec:topk}

\begin{figure}[ht]
  \centering
  \includegraphics[width=0.85\linewidth]{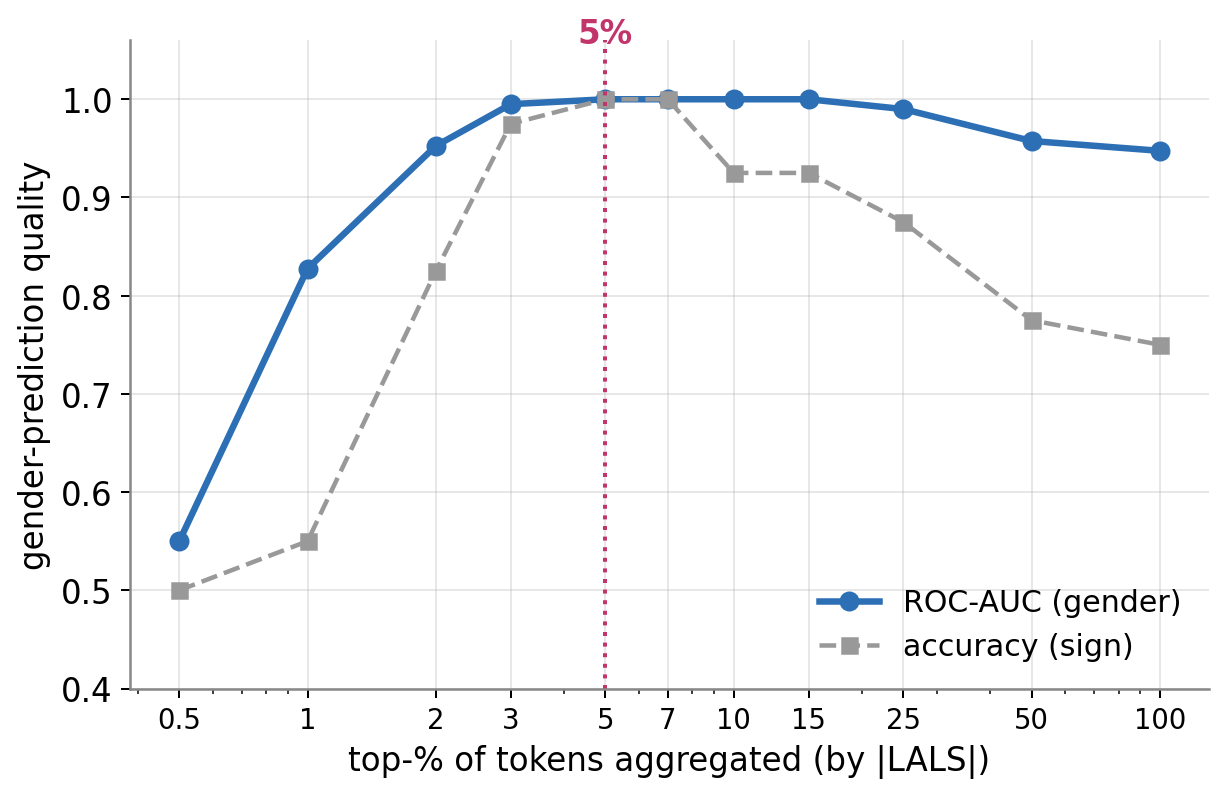}
  \caption{\textbf{Top-\% token aggregation ablation} (Qwen2-VL, visible-gender held-out set). ROC-AUC for gender prediction (solid) and sign accuracy (dashed) versus top-\% of tokens aggregated by $|$\lals{}$|$.}
  \label{fig:top_pct_ablation}
\end{figure}

Figure~\ref{fig:top_pct_ablation} validates the choice of top-5\% aggregation used throughout the paper. We compute \lals{} on a held-out visible-gender set (Qwen2-VL) and measure two metrics as a function of the top-\% of tokens aggregated by $|$\lals{}$|$: ROC-AUC for predicting visible gender, and sign accuracy (whether the image-level \lals{} matches the true gender). Both metrics peak between 5--7\% and degrade as low-magnitude tokens dilute the signal. We adopt 5\% as the default, but the qualitative findings are stable across the 3--15\% range.

\subsection{Gender Lexicons}
\label{sec:lexicon}
Table~\ref{tab:lexicon} lists the main reference lexicon used for all \lals{}
results in the paper. It contains female-associated and male-associated
terms---role nouns, pronouns, adjectives, and their plurals---balanced in
size across the two poles. To check that \lals{} does not hinge on this
particular word choice, we repeat the analysis with two alternative lexicons
(Table~\ref{tab:lexicon-controls}): \textsc{Lexicon-Roles}, restricted to
kinship/role nouns and pronouns, and \textsc{Lexicon-Names}, consisting only
of gendered first names with no role or attribute words at all.

\begin{table}[t]
    \centering
    \small
    \begin{tabular}{@{}p{0.46\linewidth}p{0.46\linewidth}@{}}
        \toprule
        \textbf{Female} (21 terms) & \textbf{Male} (21 terms) \\
        \midrule
        woman, girl, female, she, her, mother, wife, daughter, sister, lady,
        feminine, pregnant, actress, heroine, women, girls, ladies, mothers,
        wives, daughters, sisters &
        man, boy, male, he, him, his, father, husband, son, brother,
        gentleman, masculine, actor, hero, men, boys, gentlemen, fathers,
        husbands, sons, brothers \\
        \bottomrule
    \end{tabular}
    \caption{\textbf{Representative main gender reference lexicon}}
    \label{tab:lexicon}
\end{table}

\begin{table*}[t]
    \centering
    \small
    \begin{tabular}{@{}p{0.48\textwidth}p{0.48\textwidth}@{}}
        \toprule
        \textbf{Female} & \textbf{Male} \\
        \midrule
        \multicolumn{2}{@{}l@{}}{\textsc{Lexicon-Roles} --- sensitivity
        control, role nouns and pronouns only (28 / 28 terms)} \\
        \addlinespace[2pt]
        woman, women, female, females, lady, ladies, girl, girls, mother,
        mothers, mom, moms, mommy, sister, sisters, daughter, daughters, wife,
        wives, aunt, aunts, grandmother, grandmothers, grandma, she, her, hers,
        herself &
        man, men, male, males, gentleman, gentlemen, boy, boys, father,
        fathers, dad, dads, daddy, brother, brothers, son, sons, husband,
        husbands, uncle, uncles, grandfather, grandfathers, grandpa, he, him,
        his, himself \\
        \midrule
        \multicolumn{2}{@{}l@{}}{\textsc{Lexicon-Names} --- sensitivity
        control, first names only (101 / 102 terms)} \\
        \addlinespace[2pt]
        mary, patricia, jennifer, linda, elizabeth, barbara, susan, jessica,
        sarah, karen, lisa, nancy, betty, sandra, margaret, ashley, kimberly,
        emily, donna, michelle, carol, amanda, melissa, deborah, stephanie,
        rebecca, laura, sharon, cynthia, kathleen, amy, shirley, angela,
        helen, anna, brenda, pamela, nicole, samantha, katherine, christine,
        emma, ruth, catherine, debra, rachel, carolyn, janet, virginia, maria,
        heather, diane, julie, joyce, victoria, kelly, christina, joan,
        evelyn, lauren, judith, olivia, frances, martha, cheryl, megan,
        andrea, hannah, jacqueline, ann, gloria, jean, kathryn, alice, teresa,
        sara, janice, marie, julia, grace, judy, theresa, madison, beverly,
        denise, marilyn, amber, danielle, abigail, brittany, rose, diana,
        natalie, sophia, alexis, lori, kayla, jane, alyssa, tiffany,
        isabella &
        james, john, robert, michael, william, david, richard, joseph, thomas,
        charles, christopher, daniel, matthew, anthony, donald, mark, paul,
        steven, andrew, kenneth, george, joshua, kevin, brian, edward, ronald,
        timothy, jason, jeffrey, ryan, jacob, gary, nicholas, eric, jonathan,
        stephen, larry, justin, scott, brandon, frank, benjamin, gregory,
        samuel, raymond, patrick, alexander, jack, dennis, jerry, tyler,
        aaron, henry, jose, adam, douglas, nathan, peter, zachary, kyle,
        walter, harold, jeremy, ethan, carl, keith, roger, gerald, christian,
        terry, sean, arthur, austin, noah, lawrence, jesse, joe, bryan, billy,
        jordan, albert, dylan, bruce, willie, gabriel, alan, juan, logan,
        wayne, ralph, roy, eugene, randy, vincent, russell, louis, philip,
        bobby, johnny, bradley, elijah, mason \\
        \bottomrule
    \end{tabular}
    \caption{\textbf{Reference lexicons.}}
    \label{tab:lexicon-controls}
\end{table*}

\section{Robustness Checks}
\label{sec:robustness}

\subsection{Localization Replicates Across Scene Types}

\begin{figure}[ht]
\centering
\includegraphics[width=0.85\linewidth]{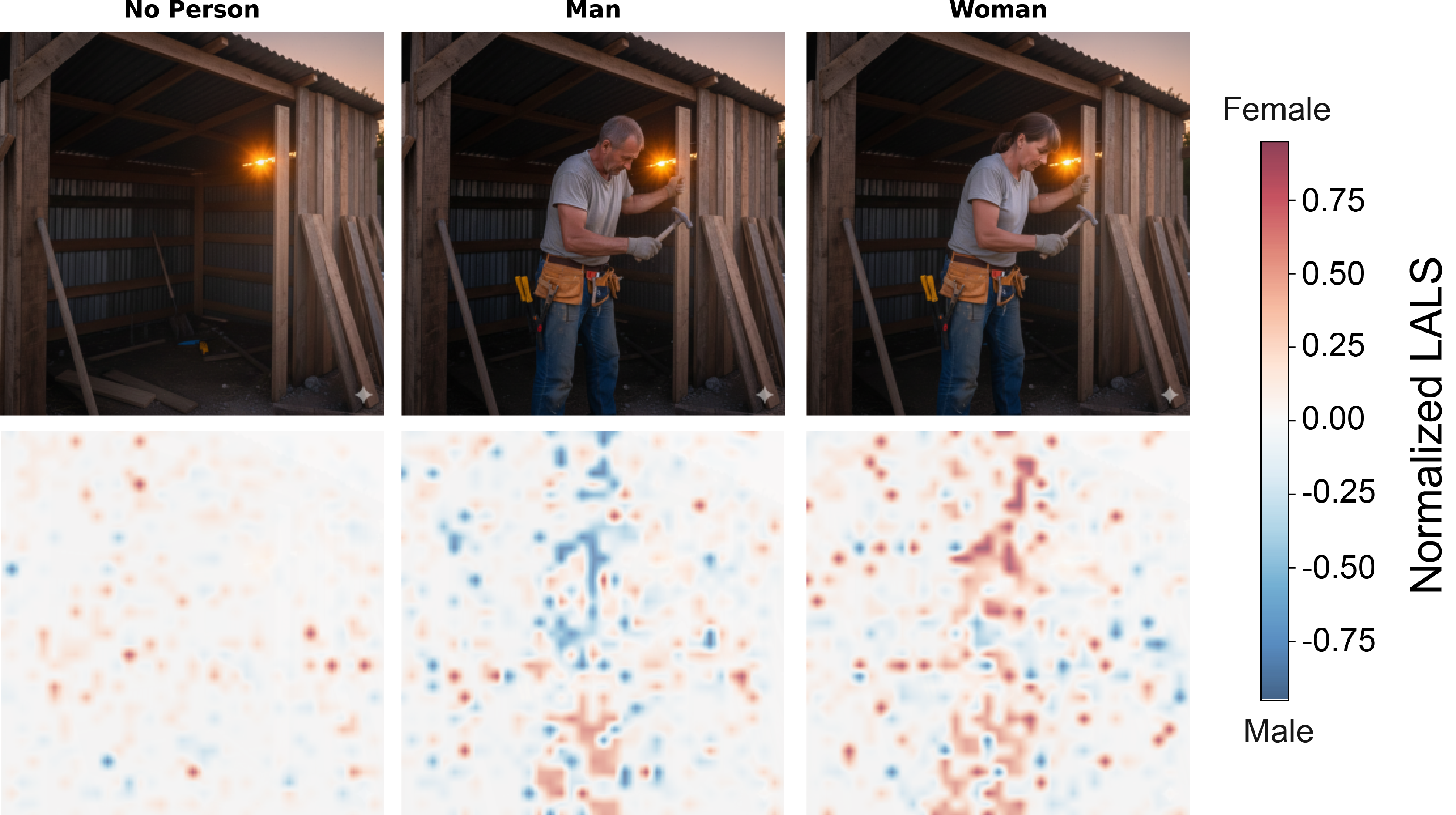}
\caption{\textbf{Construction-site replication.} The empty scene is neutral; inserting a man shifts the signal toward male (blue) and inserting a woman shifts it toward female (red), confirming that the kitchen-scene result generalizes across scene types.}
\label{fig:working_panel}
\end{figure}

Figure~\ref{fig:working_panel} reproduces the kitchen-scene gender localization experiment from Section~4.1 in a construction-site setting. The empty scene yields near-zero \lals{}; inserting a man shifts the signal toward male (blue) and inserting a woman shifts it toward female (red), with the response localized on the inserted figure. This confirms that \lals{} responds to gender cues in the image rather than to scene context.

\subsection{Real Photographs vs.\ Synthetic Images}

\begin{figure}[ht]
  \centering
  \includegraphics[width=\linewidth]{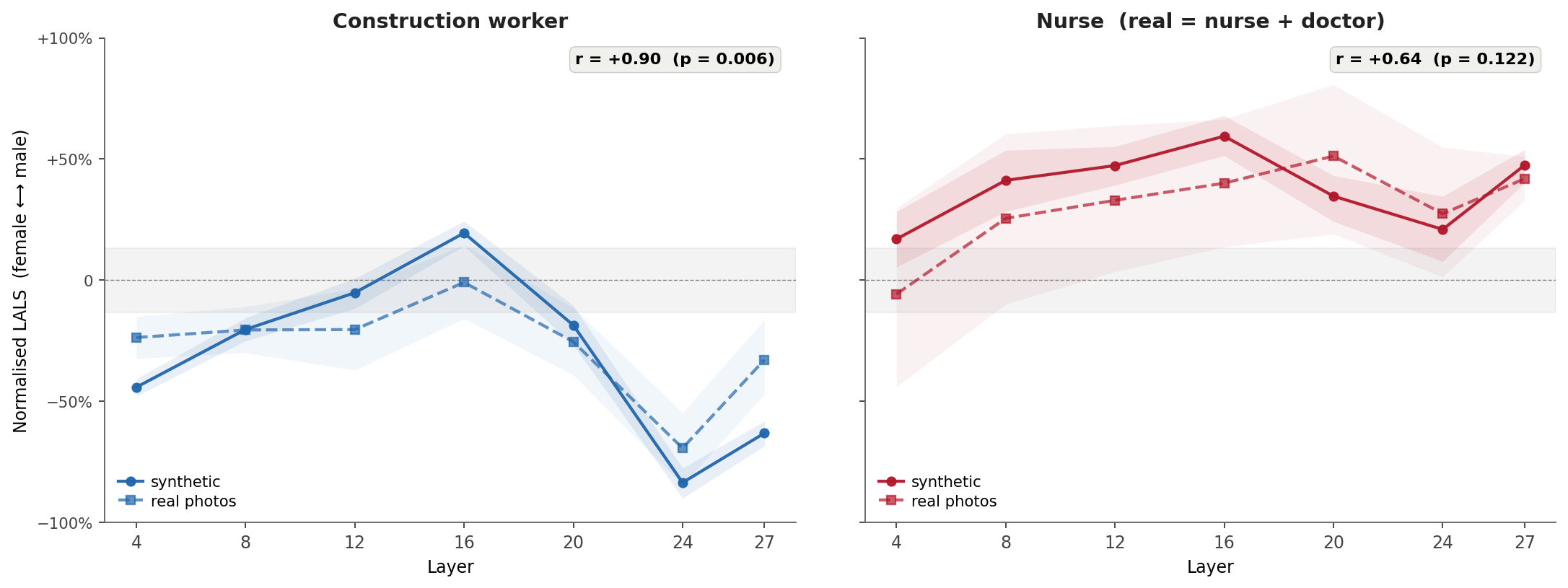}
  \caption{\textbf{Real vs.\ synthetic images} (Qwen2-VL, $N{=}10$/condition; mean $\pm$ s.e.m.). Layer-wise \lals{} on real photographs follows the same trajectory as on synthetic images. Shaded band: neutral zone.}
  \label{fig:realimages}
\end{figure}

A natural concern is that our findings may be specific to AI-generated images. Figure~\ref{fig:realimages} compares layer-wise \lals{} trajectories on real photographs to those on our synthetic dataset for construction workers and nurses (Qwen2-VL, $N{=}10$ per condition). The trajectories are closely aligned: Pearson $r{=}0.90$ ($p{=}0.006$) for construction workers and $r{=}0.64$ ($p{=}0.122$) for nurses. The lower significance for nurses reflects the small sample size (the shape of the curve matches well, but with $N{=}10$ the correlation test is underpowered). The real-photo nurse set pools nurse and doctor images, both of which wear scrubs and are gender-ambiguous from typical angles.

\section{Extended Layer Analyses}
\label{sec:extended_layers}

\subsection{Per-Architecture Layer Sweep}

\begin{figure*}[ht]
  \centering
  \includegraphics[width=\linewidth]{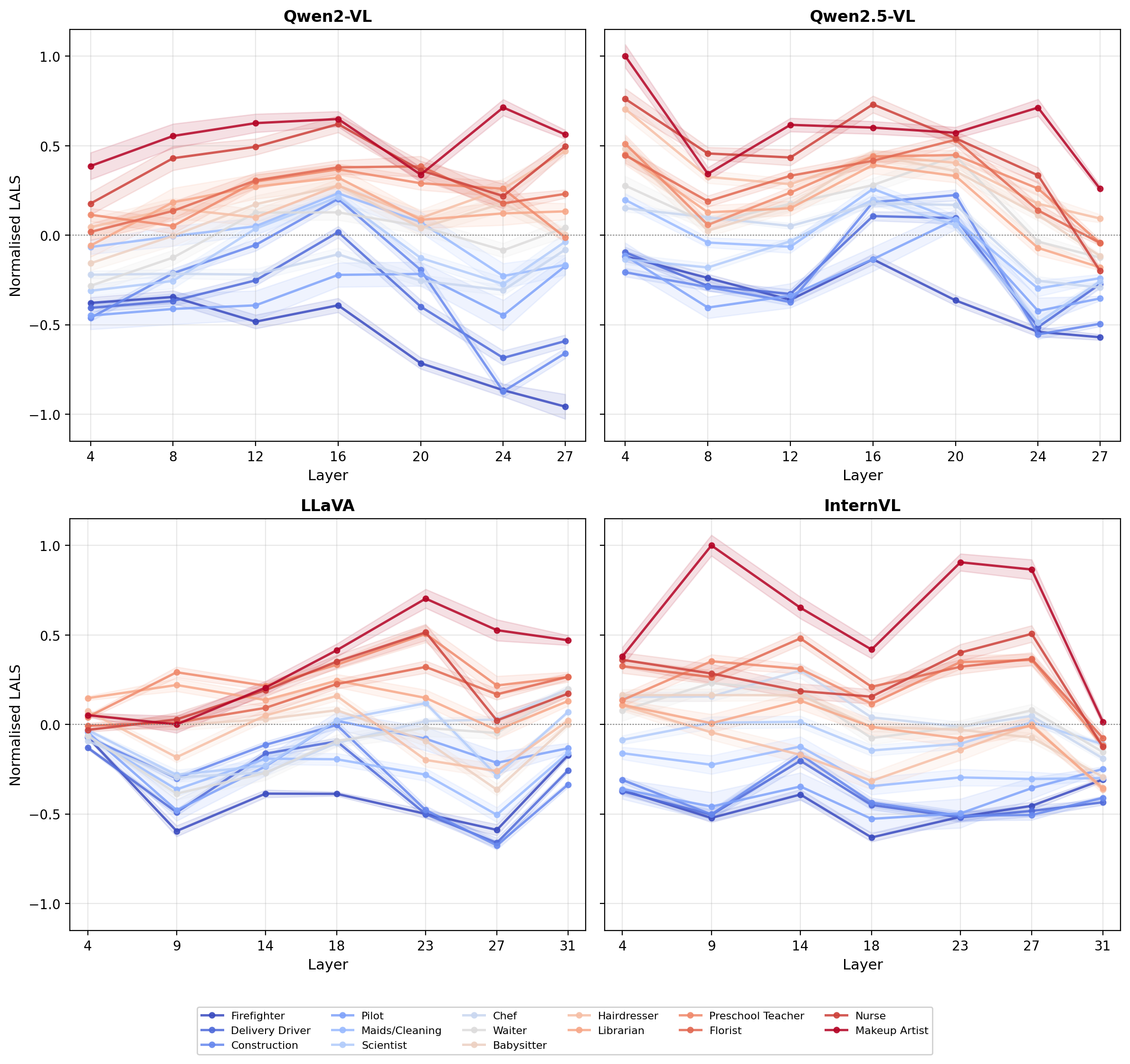}
  \caption{\textbf{Per-architecture layer sweep.} Normalised \lals{} across layers for 15 occupations and four VLM architectures ($N{=}25$ images per occupation). Each line is one occupation; blue = male-leaning, red = female-leaning.}
  \label{fig:layer_sweep}
\end{figure*}

Figure~\ref{fig:layer_sweep} shows the full per-architecture layer sweep across all 15 occupations and the four VLMs we evaluate. The qualitative pattern is consistent across architectures: male-leaning occupations enter the network with negative \lals{} and remain so through the final layer, while female-leaning occupations peak in mid-network depths (layers $\sim$12--16 for the Qwen models; $\sim$14--23 for LLaVA and InternVL) and are attenuated before the output. LLaVA exhibits the mildest collapse, compressing female signals toward zero rather than crossing into male-leaning space, while InternVL2.5 shows the strongest late-layer suppression---consistent with its near-100\% male forced-choice rates on most occupations (Table~1, main paper). Despite differences in vision encoders, vision--language connectors, and training data, the male-amplify/female-suppress asymmetry is recovered in every architecture we tested.

\subsection{Instruct vs.\ Base Model}

\begin{figure*}[ht]
  \centering
  \includegraphics[width=\linewidth]{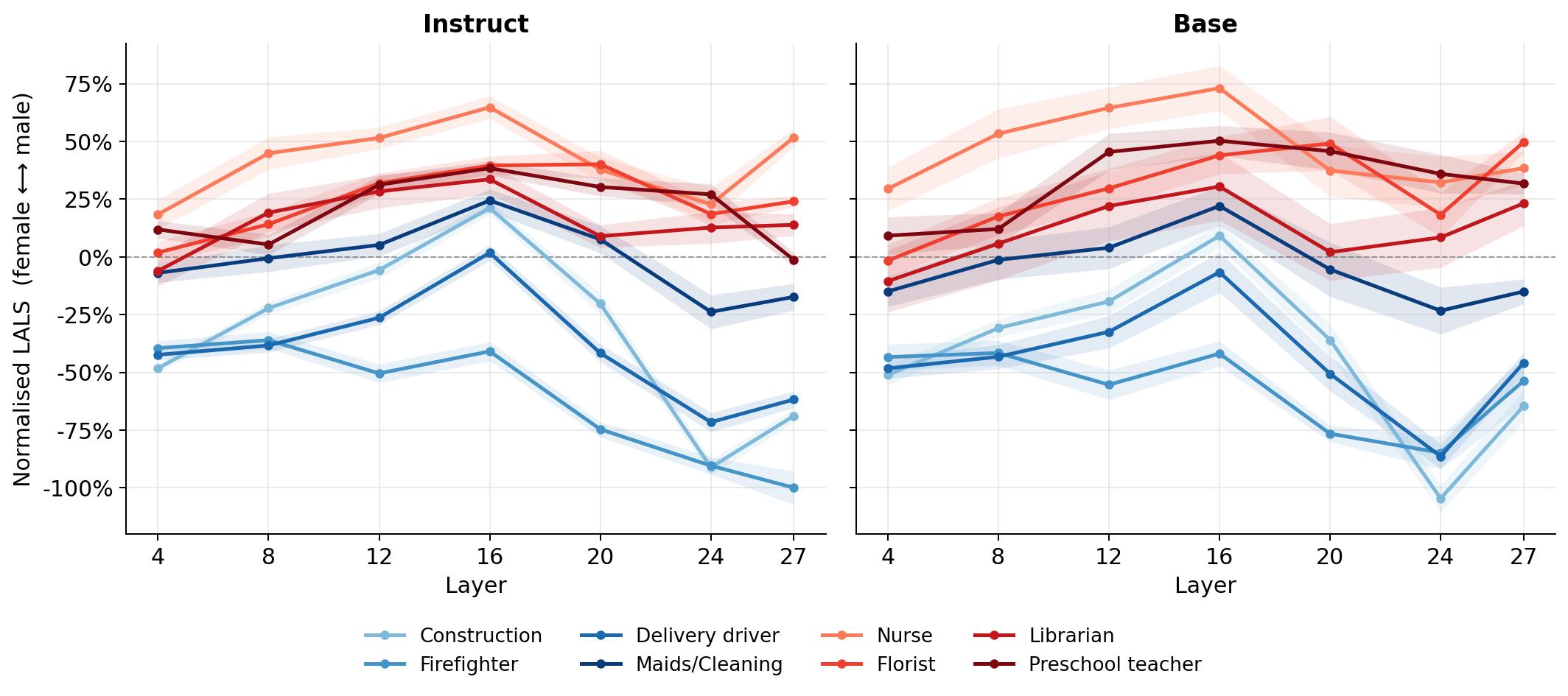}
  \caption{\textbf{Instruct vs.\ base model comparison} (Qwen2-VL-7B, $N{=}25$ images per occupation; mean $\pm$ s.e.m.). Normalised \lals{} across network depth for the instruction-tuned model (\textbf{left}) and the pre-RLHF base model (\textbf{right}). The asymmetric filtering pattern is present in both variants; instruction tuning amplifies but does not introduce it.}
  \label{fig:base_vs_instruct}
\end{figure*}

To test whether the late-layer suppression of female signal is introduced by instruction tuning, we run the same \lals{} layer sweep on the Qwen2-VL-7B base checkpoint (no RLHF) and compare it to the instruct variant (Fig.~\ref{fig:base_vs_instruct}). Both variants show the same occupation-dependent profiles: female-stereotyped occupations (red) peak around layer 16 and decline toward the output, while male-stereotyped occupations (blue) maintain or amplify their signal end-to-end. The late-layer collapse is milder in the base model—suggesting that instruction tuning amplifies the suppression—but the asymmetric structure is already present before RLHF, indicating that it is established during pretraining rather than introduced by alignment.

\subsection{Text-Only Condition}
\label{sec:textonly}
To test whether the late-layer collapse is inherited from the language backbone, we feed occupation names as text-only prompts (no image) and compute \lals{} on the resulting text tokens at each layer (Qwen2-VL-7B-Instruct). We found that for female-stereotyped occupations (nurse, florist, librarian): in the text-only condition the female signal \emph{amplifies} toward the output rather than collapsing, the opposite of the image trajectories in Fig.~\ref{fig:layer_sweep}. The late-layer collapse is therefore specific to the visual pathway.

\section{Causal Intervention: Is the Mid-Layer Signal Necessary and Sufficient?}
\label{sec:causal}

The layer sweeps in Section~\ref{sec:layer_dynamics}
show that female-associated occupations carry a mid-network \lals{} peak that is
attenuated before the final layer. These results are correlational: they
establish that the signal exists and that the model nevertheless outputs
\textit{male}, but they do not show that the mid-layer signal is part of the
causal pathway to the output. To test this, we directly ablate the signal and
measure whether the forced-choice output moves, and then run the symmetric
additive intervention to test whether adding the signal flips the output. 

\paragraph{Method.}
We perform a single-direction activation intervention on Qwen2-VL-7B-Instruct. (i)~We construct a gender direction $\mathbf{d} \in \mathbb{R}^{d}$ from the model's own text embeddings as the difference between the mean female-term and mean male-term embedding, normalised to unit length. (ii)~At layer~16 (the mid-network peak identified in Fig.~\ref{fig:layer_sweep}), we register a forward hook that projects $\mathbf{d}$ out of every visual-token hidden state during the forward pass:
\[
    \mathbf{h}_t^{(16)} \;\leftarrow\; \mathbf{h}_t^{(16)} \;-\; \alpha\,(\mathbf{h}_t^{(16)} \cdot \mathbf{d})\,\mathbf{d},
\]
with $\alpha=1$ (full ablation along $\mathbf{d}$). Text tokens are left untouched, and no other layer is modified. (iii)~We re-run the model with the hook attached and measure both \lals{} at layer~16 and the forced-choice output. We evaluate $N=20$ ambiguous images per occupation across seven occupations spanning the three regimes from Section~\ref{sec:layer_dynamics}.

\begin{figure}[t]
    \centering
    \includegraphics[width=\linewidth]{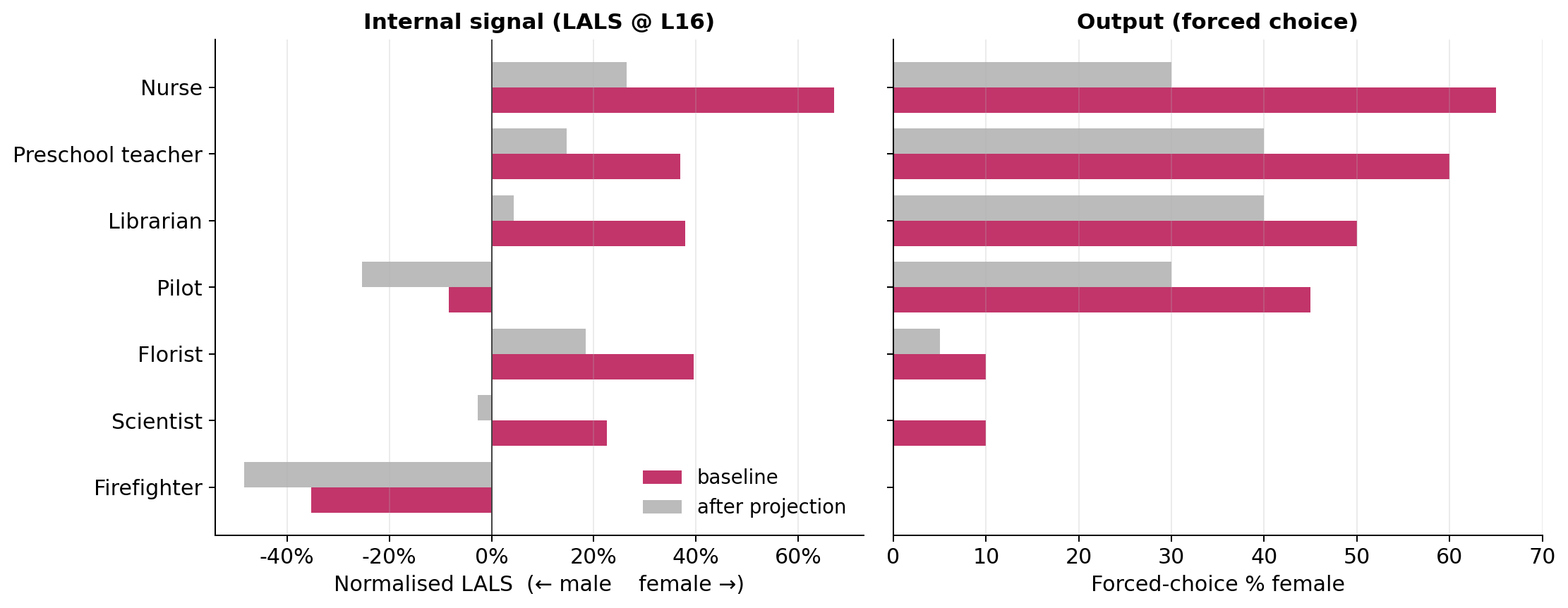}
    \caption{\textbf{Causal intervention at layer~16} (Qwen2-VL-7B-Instruct,
    $N{=}20$ images per occupation). We project a single gender direction
    out of the visual-token activations at layer~16 during the forward pass
    (full ablation, $\alpha{=}1$) and re-run the model. \textbf{Left:} the
    mid-layer \lals{} signal collapses for female-leaning occupations.
    \textbf{Right:} the forced-choice female rate drops in lockstep, while
    male-default occupations are unaffected. This indicates that the
    mid-layer signal \lals{} detects is part of the causal pathway to the
    output, not a passive correlate.}
    \label{fig:causal-intervention}
\end{figure}

\paragraph{Result.} Figure~\ref{fig:causal-intervention} shows the effect. Ablating the gender direction at layer~16 collapses the internal \lals{} signal by roughly 60--90\% for the female-leaning occupations (nurse, preschool teacher, librarian, florist), and the forced-choice female rate drops in lockstep: nurse $65 \rightarrow 30\%$, preschool teacher $60 \rightarrow 40\%$, librarian $50 \rightarrow 40\%$, pilot $45 \rightarrow 30\%$. Male-default occupations move negligibly (firefighter stays at $0\%$ female), consistent with the direction $\mathbf{d}$ being aligned with female rather than male signal.

\paragraph{Interpretation.} Two findings follow. First, the mid-layer gender association \lals{} detects is \textit{causally involved} in the model's output: removing a single direction at a single mid-network layer is enough to shift the forced-choice distribution in the predicted direction across four occupations. This rules out the alternative reading in which \lals{} picks up an epiphenomenal cluster of female-coded contextual vocabulary (scrubs, classroom) that the model ignores when generating. Second, the effect is partial---nurse drops to $30\%$ female rather than $0\%$---which is expected: a one-direction, one-layer ablation cannot eliminate gender information that is distributed across directions, tokens, and layers.

\paragraph{Additive steering.} 
To complement the ablation with the symmetric test, we add a female direction rather than removing one. We construct the direction as the unit-normalised mean difference between the visual-token activations of visibly female and visibly male reference images, add it, scaled by a magnitude $\beta$, to the visual-token hidden states of the ambiguous images during the forward pass, and read off the forced-choice answer. We include two controls: a random unit vector of the same norm injected at the same layers and magnitudes, and a sweep over early, middle and late layers. We evaluate three divergence occupations (majority-male output despite female-leaning mid-layer \lals{}) with $N{=}30$ ambiguous images per occupation, on Qwen2-VL-7B-Instruct and LLaVA-v1.6-Mistral-7B. For Qwen2-VL, adding the female direction at the middle layers moves the forced-choice output from roughly $22\%$ female at baseline to $100\%$ female as $\beta$ grows, averaged over the three occupations. For LLaVA the effect is sharper, reaching about $97\%$ female at a moderate magnitude before the injection becomes large enough that generation degrades. The matched random vector reproduces none of this, staying within a few points of baseline at every layer and magnitude, and injecting at the late layers at the same magnitude does not move the output. The two interventions are thus the two halves of the causal validation: removing the mid-layer direction suppresses female answers, adding it produces them, and an equally large random perturbation does neither, which rules out the reading in which the ablation merely damaged the representation.

\paragraph{Scope and caveats.}
The experiments are deliberately narrow and we flag the corresponding limits: 
(i)~\textit{Two constructions of the direction.} The ablation removes a 
text-derived direction while the steering adds an image-derived one; the two
are constructed independently and we do not claim they are identical. We make
no debiasing or correction claim from either: steering that drives the female
rate to $100\%$ replaces one default with another.
(ii)~\textit{Not a localisation of the
late-layer suppression.} The ablation hook is placed at layer~16, where the female 
signal peaks; the experiments test that this signal matters for the output, 
not where in the late layers it is filtered out. Identifying the specific
components responsible for the late-layer collapse---attention heads,
residual-stream subspaces, or specific MLP blocks---is an open question.
(iii)~\textit{Coverage.} The ablation is reported for Qwen2-VL-7B-Instruct 
at layer~16 with $\alpha{=}1$, and steering for two of the four
architectures; whether either direction transfers to Qwen2.5-VL and
InternVL2.5, or whether intermediate $\alpha$ values trace a smooth
dose-response curve for the ablation, is left to future work.
(iv)~\textit{Sample size.} $N{=}20$ (ablation) and $N{=}30$ (steering) per 
occupation are sufficient to observe the qualitative shifts but not to
support fine-grained per-occupation significance claims; we report both as
confirmatory mechanistic checks rather than quantitative effect estimates.
(v)~\textit{Matched-magnitude late-layer null.} Residual-stream norms are 
several times larger at the late layers than at the middle layers, so the
same absolute $\beta$ is a proportionally smaller perturbation there; the
late-layer result should be read with that in mind rather than as a strict
claim.

\section{Preliminary Retrieval Study: What Embedding-Based Systems Inherit}
\label{sec:retrieval}
 
Our results motivate representation-level auditing with the reuse of VLM embeddings in retrieval and ranking systems, where representations act without passing through the language head. As a first test of this pathway, we run a nearest-neighbour retrieval experiment on the visual embeddings alone. We assemble a small gallery of clearly gendered reference images and, for each ambiguous occupation image, retrieve its nearest neighbours in embedding space and read off their gender, after accounting for the strong anisotropy of the embedding space. The direction of the result is consistent with Section~\ref{sec:layer_dynamics}: ambiguous images of divergence occupations sit closest to female references (florist and preschool teacher retrieve roughly 70--80\% female neighbours), while a male anchor such as construction worker stays near 0\% female. What a downstream embedding-based system inherits therefore appears to track the internal representation rather than the male forced-choice output.
 
Two caveats prevent us from making a downstream claim on this basis. First, the embeddings are highly anisotropic, and nearest-neighbour distance in this space may not be a reliable readout of gender geometry even after correction. Second, the reference gallery is small and ties gender to scene: the female references are indoor service settings and the male references outdoor physical ones, so part of the apparent female association could reflect scene similarity rather than gender. We therefore report this result as preliminary.

\section{Reproducibility: Licenses and Compute} \label{sec:reproducibility} 
\paragraph{Model licenses.} All four VLMs we evaluate are open-weight and used strictly for forward-pass inference and activation extraction (no fine-tuning, no weight redistribution). Qwen2-VL-7B (both instruct and base), Qwen2.5-VL-7B, and LLaVA-v1.6-Mistral-7B are released under the Apache~2.0 license; InternVL2.5-8B is released under the MIT license. The ambiguous-occupation images were generated with Google Gemini~2.5 Flash under its standard terms of service; the dataset is fully synthetic and contains no real individuals. 

\paragraph{Compute.} All experiments ran on a single NVIDIA H100 (80\,GB) GPU and consumed approximately 25 GPU-hours in total, covering activation extraction, forced-choice and chain-of-thought generation across the four models, the robustness checks, the color ablation, the causal intervention and other experiments. \lals{} itself is training-free and adds negligible overhead beyond the forward pass.

\end{document}